\documentclass{article}

\usepackage{microtype}
\usepackage{graphicx}
\usepackage{subcaption}
\usepackage{booktabs} % for professional tables
\usepackage{float}
\usepackage{afterpage}

\usepackage{hyperref}

\usepackage[accepted]{icml2026}

\usepackage{amsmath}
\usepackage{amssymb}
\usepackage{mathtools}
\usepackage{amsthm}
\usepackage{multirow}
\usepackage{multicol}

\usepackage[capitalize,noabbrev]{cleveref}

\theoremstyle{plain}

\theoremstyle{definition}

\theoremstyle{remark}

\usepackage[textsize=tiny]{todonotes}

\icmltitlerunning{PGD-NO: A Neural Operator with Precomputed Geometry Decomposition}

\begin{document}

\twocolumn[
  \icmltitle{PGD-NO: A Neural Operator with Precomputed Geometry Decomposition \\  for 3D Million-scale Physics Simulations}



  \icmlsetsymbol{equal}{*}

  \begin{icmlauthorlist}
    \icmlauthor{Weiheng Zhong}{ceeuiuc,ds}
    \icmlauthor{Jing Bi}{ds}
    \icmlauthor{Victor Oancea}{ds}
    \icmlauthor{Hadi Meidani}{ceeuiuc,sscds}
  \end{icmlauthorlist}

  \icmlaffiliation{ceeuiuc}{Department of Civil and Environmental Engineering, Grainger College of Engineering, University of Illinois Urbana-Champaign, Urbana, IL, US}
  \icmlaffiliation{sscds}{Siebel School of Computing and Data Science, Grainger College of Engineering, University of Illinois Urbana-Champaign, Urbana, IL, US}
  \icmlaffiliation{ds}{Dassault Systemes Americas, Johnston, RI, US}
  
  \icmlcorrespondingauthor{Weiheng Zhong}{weiheng4@illinois.edu}

  \icmlkeywords{Neural Operator, computational physics, industry-scale}

  \vskip 0.3in
]



\printAffiliationsAndNotice{}  % no special notice (required even if empty)

\begin{abstract}
  While neural PDE solvers have demonstrated significant potential for accelerating engineering simulations, existing architectures remain constrained by high memory consumption and the "single-node bottleneck," where the maximum processable mesh resolution is strictly limited by the VRAM of a single compute unit. To address these challenges, we propose PGD-NO, a neural operator with Precomputed Geometry Decomposition, that relocates the computational overhead of geometric encoding to a deterministic pre-computation phase. By utilizing an iterative geometry decomposition algorithm to extract "geometry tokens," our model decouples feature extraction from solution querying. This architecture enables linear memory scalability, allowing high-fidelity learning on meshes exceeding 10 million nodes—a scale where existing architectures typically encounter memory exhaustion. PGD-NO demonstrates competitive predictive accuracy across diverse industrial benchmarks and provides intrinsic interpretability through attention mechanisms. By effectively overcoming traditional mesh-size constraints, PGD-NO offers a robust and efficient solution for the next generation of large-scale, high-fidelity industrial design applications. Our codes and datasets are available at \href{https://github.com/WeihengZ/PGD-NO}{https://github.com/WeihengZ/PGD-NO}. 
\end{abstract}

\section{Introduction}

The numerical solution of partial differential equations (PDEs) is a cornerstone of computational science and engineering, facilitating evaluations \cite{fem}, sensitivity analysis \cite{fem_sensitivity}, and uncertainty quantification \cite{pivae} in industrial design. Traditional discretization-based methods—such as the finite element method (FEM) \cite{fem} and finite volume method (FVM) \cite{fvm} —are highly refined, offering significant accuracy and robustness across diverse domains. Nevertheless, these conventional approaches often involve prohibitive computational costs, particularly for large-scale, high-fidelity simulations \cite{DON}. To mitigate these demands, deep learning architectures have been investigated as efficient surrogates for numerical methods, called neural PDE solvers \cite{lu2021, FNO, transolver}. By training on simulation data, these neural solvers learn to approximate the mapping between inputs and outputs of the traditional methods, thereby significantly accelerating the manufacturing and design process.

Industrial applications typically involve large, complex geometries, requiring models to efficiently capture the underlying physical features \cite{transolver}. DeepONet \cite{lu2021} was initially applied to industrial design because many geometries are governed by low-dimensional parameter vectors for manufacturing purposes. However, recent studies indicate that the performance of DeepONet diminishes as geometric complexity increases \cite{FNO}. Simultaneously, various deep learning models have been proposed for freeform geometries. Although these methods show promise in benchmark problems, they struggle with real-world industrial tasks involving millions of mesh points \cite{transolver_plus}. This is because training on such massive datasets requires significant GPU memory, which often exceeds the capacity of standard hardware.

\begin{figure*}[t]
  \begin{center}
    \centerline{\includegraphics[width=2.0\columnwidth]{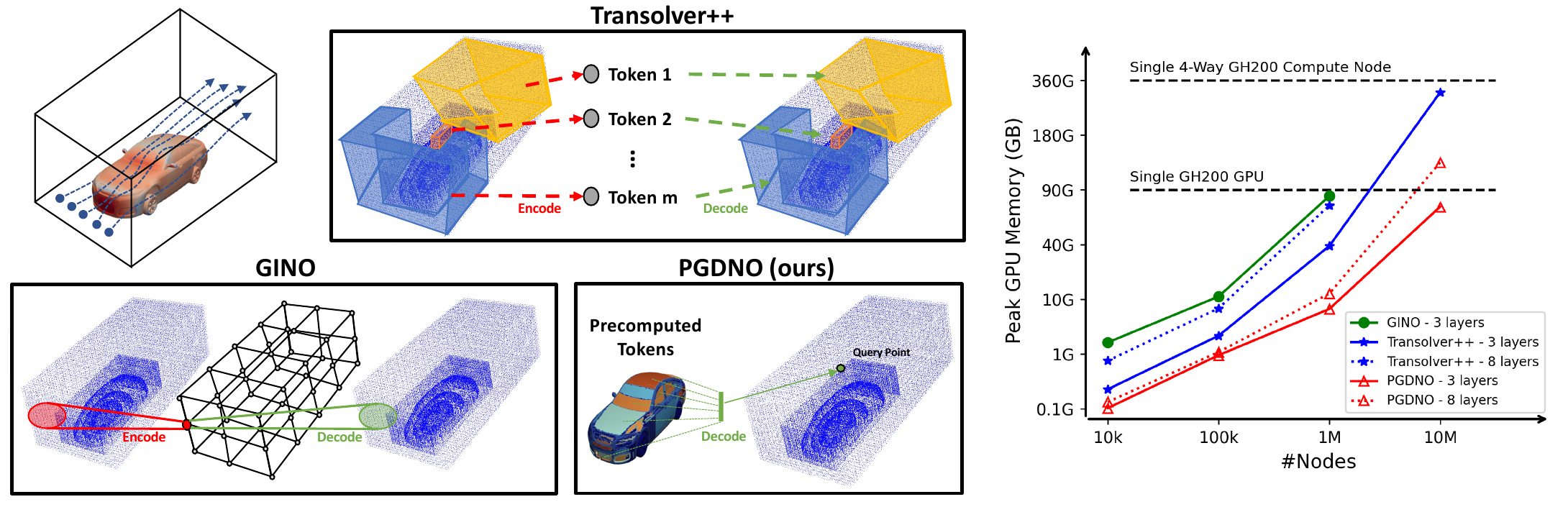}}
    \caption{Architectural comparison and memory scaling. (Left) Encoding mechanisms: GINO utilizes grid-based mapping, and Transolver++ computes physics tokens via weighted nodal sums, whereas PGD-NO employs pre-computed geometry tokens to bypass intensive encoding. (Right) Peak GPU memory on a GH200 system: PGD-NO maintains the lowest footprint, enabling high-resolution simulations (up to 10M nodes) that exceed the capacity of traditional grid-based or graph-based neural operators.}
    \label{fig.sellpoint}
  \end{center}
\end{figure*}

Existing neural PDE solvers for freeform geometries can be categorized into graph-based, grid-based, and point-based models, depending on how they represent PDE solutions on a mesh. Graph-based models \cite{GNO, velasco2024gtnn, li2020multipole, EAGNO} may fail to capture complex geometries if they rely on sparse mesh graphs; conversely, they can rapidly exhaust GPU memory if they utilize radius-based methods to synthesize edges. Grid-based models \cite{hashgrid}, such as GINO \cite{li2023gino}, are also memory-intensive, as capturing complex geometries requires mapping point features onto high-resolution grids. In contrast, point-based methods \cite{pointnet, Park2025, lee2024inducing}, such as Transolver \cite{transolver}, utilize a physics-attention mechanism to identify underlying "physics tokens," demonstrating that complex geometric features can be captured directly from point clouds. Recently, Transolver++ \cite{transolver_plus} has extended this capability to million-scale problems through an optimized parallelism framework.

Despite its advancements, Transolver++ still faces limitations when scaled to high-fidelity PDE solving tasks. In practice, we observe significant bottlenecks in training efficiency when mesh sizes exceed 10 million nodes. While Transolver++ can theoretically scale via parallel computation, this comes at the cost of increased communication overhead. Although intra-node communication remains efficient due to NVLink \cite{nvlink} architecture, scaling beyond a single compute node necessitates data transfers between compute nodes via standard networking. This inter-node communication introduces substantial latency, significantly increasing the duration of each training epoch. Consequently, the practical mesh-size capacity of Transolver++ remains constrained by the resources of a single compute node. As illustrated in Figure~\ref{fig.sellpoint}, even utilizing the latest 4-way GH200 architecture with 360GB of total memory, a Transolver++ model with a reduced number of layers can barely accommodate 10M nodes for CFD simulation learning.

We consider that the GPU memory limitations of existing methods stem from their reliance on learning-based encoding process to map complex geometries into hidden representations. In contrast, geometric features can be pre-computed using deterministic algorithms, which significantly reduces GPU memory consumption by bypassing the encoding process. Motivated by this observation, this work explores an alternative paradigm for developing neural PDE solvers. We introduce a geometry decomposition algorithm for feature extraction and propose a simple but effective decoding architecture to predict PDE solutions based on these pre-computed features. Hence, our primary contributions in this work are as follows:
\begin{itemize}
    \item We introduce a deterministic hierarchical decomposition algorithm to extract ``geometry tokens,'' demonstrating that PDE solutions can be effectively learned using non-learnable geometric primitives.
    \item We propose the Pre-computed Geometry Decomposition Neural Operator (PGD-NO), which decouples geometry encoding from solution querying, eliminating inter-node communication for million-scale mesh processing on standard hardware.
    \item We evaluate our model against current state-of-the-art methods across five industrial benchmarks, including a novel, self-generated dataset of up to 12 million nodes, where our approach achieves superior performance.
    \item We further demonstrate that PGD-NO enables efficient training via random sampling while providing enhanced interpretability for its physical predictions
\end{itemize}

\textbf{Conflict of Interest Disclosure} The authors J.B. and V.O. are employed by Dassault Systèmes, and W.Z. is currently employed by Dassault Systèmes. This paper evaluates multiple datasets, including one provided by Dassault Systèmes.

\section{Related Works}

\subsection{Neural PDE Solver}

Inspired by DeepONet, branch-trunk operators leverages the universal operator approximation theorem to map infinite-dimensional function spaces \cite{DON}. These architectures consist of a branch network encoding input functions and a trunk network processing output coordinates \cite{lu2021}. While extensions have addressed the curse of dimensionality \cite{Mandl2024}, uncertainty quantification \cite{lone2026alpha}, and computational efficiency \cite{Zhang2025, derivative_don}, and specialized variants like GANO \cite{GANO} and Point-DeepONet \cite{Park2025} incorporate geometric information, these models generally rely on fixed sensor locations or global pooling. Consequently, despite improvements in expressiveness \cite{DCON} and temporal encoding \cite{S-DON}, branch-trunk architectures remain fundamentally limited in their ability to handle truly freeform, complex geometries.

Graph-based neural operators (GNOs) \cite{GNO} are designed for irregular domains by utilizing message-passing frameworks. Key developments such as MGNO \cite{li2020multipole} and LNO \cite{lno2024} enable the capture of long-range interactions and expressive latent mappings. To generalize across arbitrary geometries, models like CORAL \cite{Neural_field}, EA-GNN \cite{EAGNO}, and PDE-GCN \cite{eliasof2021pde} have introduced coordinate-based features and PDE-aware message passing. However, a significant drawback of these methods is the memory overhead required to store and process edge information; as the mesh density increases, the memory cost of the graph connectivity can quickly exceed the capacity of modern hardware.

Grid-based operators, primarily led by the Fourier Neural Operator (FNO) \cite{FNO}, capture global dependencies in the frequency domain with high efficiency. Scalable variants like D-FNO \cite{D-FNO2025}, alongside geometry-aware models like GINO \cite{li2023gino} and DAFNO \cite{DAFNO2023}, attempt to adapt spectral methods to irregular topologies. While these models provide significant speedups \cite{BenchmarksFNO2021} and incorporate physics-informed constraints \cite{PINO2021}, they rely on mapping unstructured data to latent regular grids. In large-scale industrial applications, these grids must maintain an extremely high resolution to avoid significant information loss, which becomes computationally prohibitive.

Point-based operators offer the greatest flexibility for nonuniform resolutions and complex geometries by operating directly on spatial point clouds. Architectures such as GNOT \cite{GNOT} and Transolver \cite{transolver} utilize physics-aware attention to identify underlying tokens, while frameworks like Universal Physics Transformers (UPT) \cite{alkin2024universal} and GFN \cite{morrison2024gfn} ensure resolution invariance. Although these methods mitigate numerical artifacts \cite{zheng2024alias, yu2024nonlocal} and achieve better scalability \cite{li2023scalable}, a clear limitation persists: while their memory complexity scales linearly with the number of nodes, the maximum node capacity remains strictly capped by the GPU memory limits of a single compute node.

\subsection{Geometry Decomposition}

Traditional geometric decomposition methods \cite{tradition_geo_decomp} often rely on deterministic algorithms such as K-means clustering on mesh faces \cite{mesh_based_decomp}, hierarchical clustering based on curvature, or spectral clustering utilizing the Laplace-Beltrami operator \cite{laplace_geo_decom} to identify salient features. Advanced techniques such as convex decomposition \cite{convex_geo_decomp} aim to break complex, non-convex volumes into a set of simpler, nearly convex parts to facilitate more efficient physical simulations. In the context of industrial design, primitive-based segmentation is frequently employed to decompose geometries into standard shapes (e.g., planes, cylinders, or spheres) that correspond to manufacturing parameters \cite{drivAer, transolver_plus}. While these classical segmentation methods offer high interpretability without requiring training data, their integration into end-to-end neural PDE solvers remains largely unexplored. Furthermore, achieving further reductions in computational complexity requires a method to identify a sparse yet representative set of geometry tokens—a capability currently absent in existing geometric segmentation algorithms

\section{Method}

\subsection{Geometry Token Extraction}

The proposed workflow transforms a raw 3D mesh into a set of discrete, semantically meaningful physics tokens by combining graph topology with geometric feature detection. 

{\bfseries{Step (1) Graph Construction:}} The process begins by transforming the raw 3D mesh into a structured topological network $G = (V, E)$, where $V$ is the set of mesh vertices and $E$ is the set of edges. 

{\bfseries{Step (2) Sharp Edge Identification:}} To determine where a mesh should naturally break, the algorithm identifies sharp features using geometric gradients. The procedure evaluates the dihedral angle $\theta_{ij}$ between the normals $\mathbf{n}_a$ and $\mathbf{n}_b$ of adjacent faces sharing edge $e_{ij}$. If the angle satisfies: $\theta_{ij} = \arccos(\mathbf{n}_a \cdot \mathbf{n}_b) > \theta_{\text{threshold}}$,
the nodes $v \in e_{ij}$ are added to a node set of sharp edges $V_{\text{sharp}}$, where $\theta_{\text{threshold}}$ is the angle threshold. This acts as a structural shattering mechanism \cite{shatter}, where prominent ridges are marked for disconnection. 

{\bfseries{Step (3) Graph Partition:}} Once the sharp edges are identified, the partition is executed by creating a subgraph $G' = G[V \setminus V_{\text{sharp}}]$, where all sharp edge boundary are removed. This operation of node removal causes the graph to fracture into a set of independent connected components $\mathcal{C} = \{C_1, C_2, \dots, C_k\}$. To achieve a hierarchical refinement of the mesh, the process is performed iteratively across a sequence of thresholds. We introduce a threshold decay mechanism, $\theta^{i+1}_{\text{threshold}} = \alpha \theta^i_{\text{threshold}}$ with a hyper-parameter $\alpha \in (0, 1)$, which progressively increases the sensitivity of sharp edge detection. This decay allows the algorithm to capture coarse structural parts at higher angles before isolating finer geometric details at lower angles. Any subgraph $C_i$ that satisfies the size constraint $|C_i| > \text{min\_size}$ is preserved, thereby preventing the generation of excessively small geometry tokens that could undermine computational efficiency.

{\bfseries{Step (4) Graph Merge:}} To prevent the creation of excessive noise, the algorithm employs a merging logic based on proximity. For a small fragment $G_{\text{small}}$ where $|V_{\text{small}}| < \text{threshold}$, the algorithm identifies the boundary set $\partial V_{\text{small}}$ and expands it by $h$ hops using the adjacency relation $\text{Adj}(v)$. The merge occurs when an expanded set $V_{\text{small}}^{(h)}$ intersects with another larger graph node set $V_{\text{large}}$. Using this strategy, small graphs are absorbed into the most relevant neighbor, ensuring that the final set of tokens remains computationally manageable.

{\bfseries{Step (5) Geometry token matrix:}} In the last step, we create the mapping matrix $S \in \mathbb{R}^{K \times N}$ representing the relationship between geometry tokens and mesh nodes, where $K$ is the number of geometry tokens and $N$ is the number of mesh vertices:
\begin{equation}
    S_{ki} = 
    \begin{cases} 
      \frac{1}{|C_k|} & \text{if } v_i \in C_k, \\
      0 & \text{otherwise}.
    \end{cases}
\end{equation}

The details of the geometry token creation algorithm is shown in Algorithm~\ref{alg:mesh_seg}. 

\begin{algorithm}[t!]
  \caption{Iterative Geometry Decomposition}
  \label{alg:mesh_seg}
  \begin{algorithmic}[1]
    \STATE {\bfseries Input:} Mesh PolyData $\mathcal{P}$, Coordinates $\mathbf{X}$, Angle thresholds $\Theta = \{\theta_1, \theta_2, \dots, \theta_m\}$, Min graph size ratio $\rho$
    
    \vspace{0.8em} % Adding the void
    \STATE \textbf{1. Graph construction:}
    \STATE Build adjacency graph $G = (V, E)$ from mesh connectivity $\mathcal{P}$
    \STATE Initialize set of subgraphs to split: $\mathcal{S} = \{G\}$
    
    \vspace{0.8em} % Adding the void
    \STATE \textbf{2. Iterative partition with sharp edges:}
    \FOR{$\theta \in \Theta$}
      \STATE $V_{sharp} = \{v \in V \mid \text{dihedral angle at } v > \theta\}$
      \STATE $\mathcal{S}_{\text{next}} = \emptyset$
      \FOR{$g \in \mathcal{S}$}
        \STATE $\{\mathcal{C}_1, \dots, \mathcal{C}_k\} = \text{Graph-Split}(g \setminus V_{sharp})$
        \STATE $\mathcal{S}_{\text{next}} = \mathcal{S}_{\text{next}} \cup \{ \mathcal{C}_i \mid |\mathcal{C}_i| > \rho \cdot |V| \}$
      \ENDFOR
      \STATE $\mathcal{S} = \mathcal{S}_{\text{next}}$
      \IF{$\mathcal{S} = \emptyset$} \STATE \textbf{break} \ENDIF
    \ENDFOR

    \vspace{0.8em} % Adding the void
    \STATE \textbf{3. Graph remerge}
    \STATE $V_{unassigned} = V \setminus \bigcup_{g \in \mathcal{S}} V(g)$
    \STATE $\mathcal{S} = \mathcal{S} \cup \text{Graph-Split}(G[V_{unassigned}])$
    
    \STATE Sort $\mathcal{S}$ by size descending: $\{\mathcal{S}_{large}, \mathcal{S}_{small}\}$
    \FOR{$g_{small} \in \mathcal{S}_{small}$}
      \STATE Perform $h$-hop BFS expansion in $G$: $V_{expanded} = \text{BFS}(V(g_{small}), h)$
      \IF{$V_{expanded} \cap V(g_{large}) \neq \emptyset$}
        \STATE Merge $V(g_{small})$ into the first intersecting $V(g_{large})$
      \ENDIF
    \ENDFOR

    \vspace{0.8em} % Adding the void
    \STATE \textbf{4. Token encoding:}
    \STATE Construct sparse matrix $\mathbf{M} \in \mathbb{R}^{|\mathcal{S}| \times |V|}$
    \STATE $\text{S}_{ij} = \begin{cases} 1/|V(\mathcal{S}_i)| & \text{if } v_j \in \mathcal{S}_i \\ 0 & \text{otherwise} \end{cases}$
    
    \vspace{0.5em}
    \STATE {\bfseries Output:} Sparse segmentation matrix $\mathbf{S}$
  \end{algorithmic}
\end{algorithm}

\subsection{Model Architecture}

Upon the generation of the sparse segmentation matrix $\mathbf{S} \in \mathbb{R}^{S \times N}$, the model integrates these precomputed geometry tokens into a deep learning framework to process the input point cloud. The input mesh, represented as a point cloud of $N$ points with spatial coordinates in $\mathbb{R}^{N \times 3}$, is first projected into a high-dimensional feature space $\mathbf{X} \in \mathbb{R}^{N \times C}$. These features are then aggregated into the token space by computing the matrix product of the segmentation matrix and the point features, $\mathbf{SX} \in \mathbb{R}^{S \times C}$. This operation effectively down-samples the dense point information into a compact latent token representation of the geometry.

\begin{figure*}[t]
  \begin{center}
    \centerline{\includegraphics[width=1.8\columnwidth]{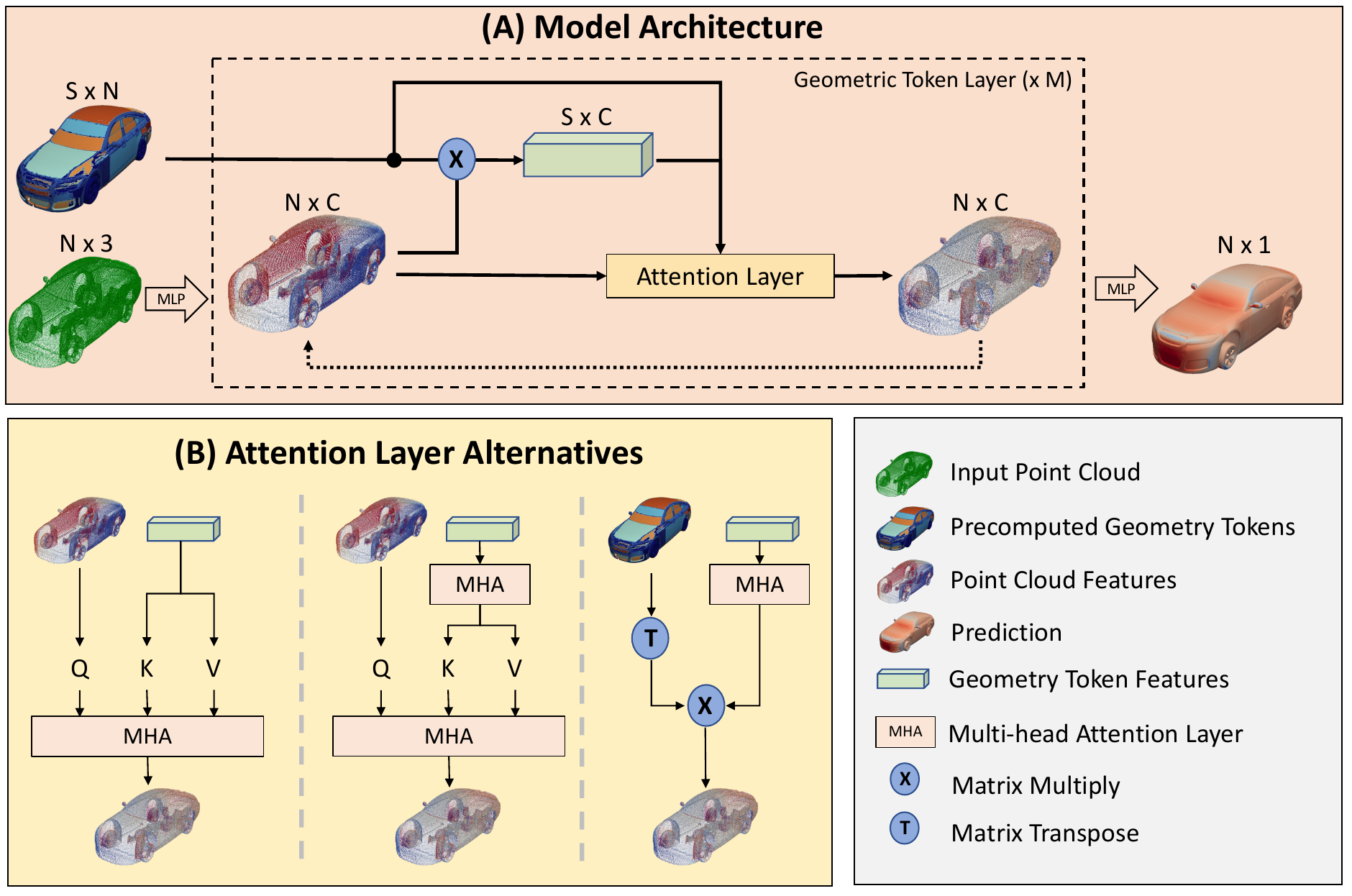}}
    \caption{PGD-NO architecture and layer variants. (Top) Model pipeline: a pre-computed segmentation matrix extracts geometry tokens from the point cloud, which are refined by stacked layers to capture multi-scale physics. (Bottom Left) Layer configurations: Variants 1 and 2 use multi-head attention to project features from tokens back to nodes, while Variant 3 uses a transposed segmentation matrix for high-efficiency surface mapping.}
    \label{fig.model_arch}
  \end{center}
\end{figure*}

The core of the architecture consists of $M$ successive Geometric Token Layers designed to refine the interaction between global geometric tokens and local point features. As shown in the model architecture diagram in Figure~\ref{fig.model_arch}, the aggregated geometry token features serve as a global context that is fed back into the nodal features with an Attention Layer, which enables the network to propagate information from semantically distinct regions back to individual points. Following the iterative refinement within the Geometric Token Layers, the final high-dimensional point features are passed through a concluding MLP to produce the target predictions.

The proposed framework introduces three architectural alternatives for the geometric token layer, varying in how they reconstruct point-wise features from the latent space. In the first variant, a multi-head cross-attention mechanism directly maps token features back to point features by utilizing the input point cloud as the Query ($Q$) and the geometry tokens as the Key ($K$) and Value ($V$). The second variant introduces an additional Multi-Head Attention (MHA) refinement step on the token features prior to the cross-attention mapping, allowing for more complex interactions within the latent representation. The third variant offers a high-efficiency alternative specifically for surface prediction by transposing the segmentation matrix to map features directly back to each mesh node. However, since these geometry tokens are precomputed based on specific mesh nodes, this third architecture is restricted to surface tasks and cannot be extended to volume node prediction. 

\subsection{Model Complexity Analysis}

To establish a baseline for performance, we evaluate the computational and memory complexity of our proposed architecture against Transolver++, which represents the best state-of-the-art neural PDE solver framework in terms of efficiency. Our analysis considers a 3D PDE problem defined by $N_s$ surface nodes, which characterize the geometry, and $N_v$ volume nodes, representing the spatial coordinates where the Partial Differential Equation (PDE) solutions are computed.

The system’s computational complexity is primarily governed by operations performed on volume node features. While our model maintains the same order of magnitude as Transolver++, it achieves a 50\% reduction in computational overhead ($O(4N_v)$ versus $O(8N_v)$) by streamlining the attention mechanism. Furthermore, by decoupling geometry extraction from solution querying, our framework significantly enhances memory efficiency. This architecture allows volume nodes to be partitioned across multiple GPUs, shifting the primary memory bottleneck to surface node operations ($O(4N_s)$), a substantial improvement over the $O(8N_v)$ bottleneck inherent in Transolver++. More details about this analysis can be found in Appendix~\ref{App.complexity_analysis}. 

\begin{figure*}[t]
  \begin{center}
    \centerline{\includegraphics[width=1.8\columnwidth]{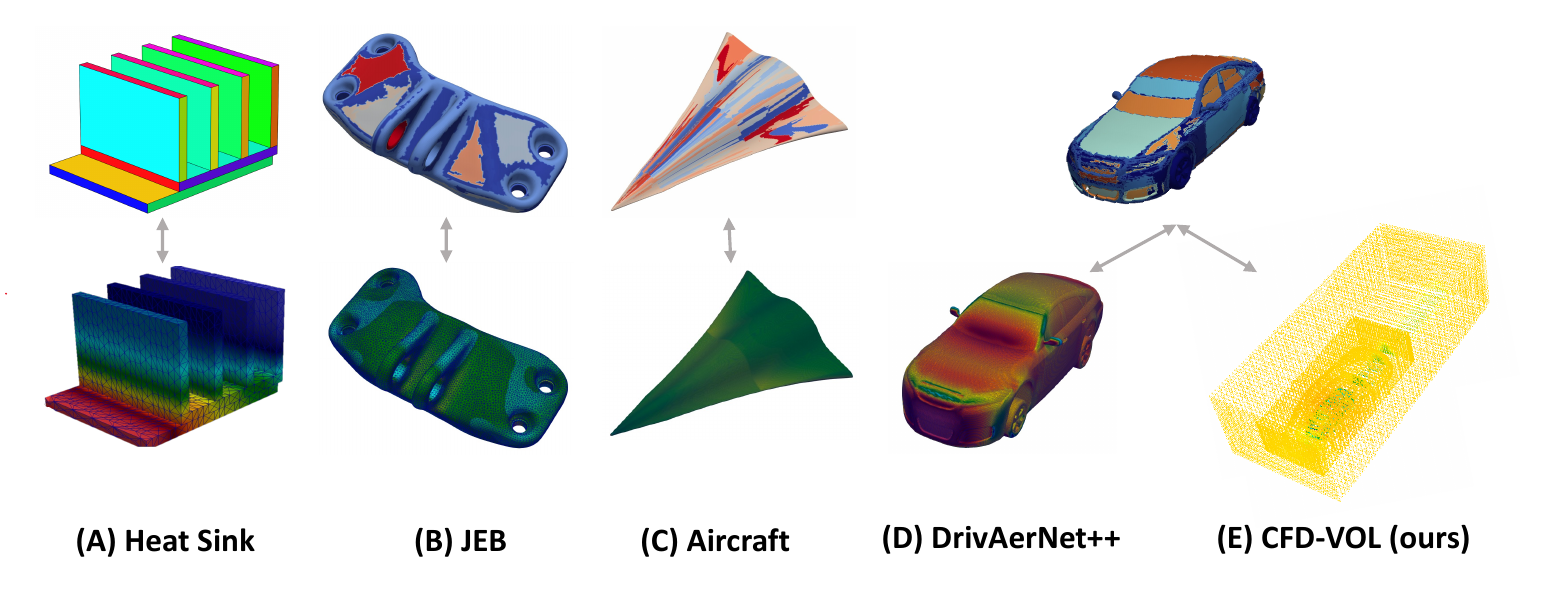}}
    \caption{Visualization of 3D industrial benchmarks. (Top) Geometry tokens generated via our iterative decomposition algorithm for representative samples. (Bottom) Corresponding ground-truth physics fields. These benchmarks cover a wide spectrum of complexities—from smooth surfaces to irregular, multi-component geometries—and diverse prediction tasks, including internal volume, surface-based, and external fluid domain patterns.}
    \label{fig.data_descrip}
  \end{center}
\end{figure*}

\section{Experiment}

\textbf{Data}: Given that two-dimensional problems have been extensively explored in the literature of Neural PDE solvers, this study focuses exclusively on assessing model performance within the context of three-dimensional, real-world industrial applications. The evaluation is conducted across the following five benchmarks: Heat Sink \cite{heatsink-dataset}, JEB \cite{deepJEB}, DrivAerNet++ \cite{drivAer++}, Aircraft \cite{transolver_plus}, and CFD-VOL proposed by us. The datasets used in this study encompass a diverse range of surface geometries, each presenting unique structural challenges for the ML model. The heat sink dataset serves as a baseline case, consisting of geometries with sharp edges formed by combinations of 3D rectangular volumes. This dataset verifies the model performance when clear geometry tokens are present, as the algorithm can easily decompose these surfaces into flat facets. In contrast, the JEB dataset consists of non-parametric geometries created by generative models. These organic shapes pose a significant challenge for the model in extracting representative geometry tokens. The aircraft dataset introduces the challenge of representing smooth surfaces that lack distinct sharp edges. Finally, the DrivAer++ dataset includes vehicle models with a high number of individual components, serving as a benchmark for the model to handle complex, multi-component assemblies. 

Prediction tasks also vary across these datasets, presenting different field-specific challenges. The heat sink and JEB datasets require the model to predict volumetric data, whereas the DrivAer++ dataset focuses on surface-based predictions. The aircraft dataset similarly involves surface prediction but adds the complexity of handling parametric controls within the physics simulation. Finally, the proposed CFD-VOL dataset presents the unique challenge of predicting physics features in the external domain surrounding the geometry. This requires the model to accurately capture fluid patterns and field interactions outside the physical boundaries of the object.

Our geometry decomposition algorithm is applied exclusively to the manifold surface. We keep all the extracted tokens of heat sink dataset and the largest $M$=128 tokens for all the other datasets. Empirical benchmarks across all datasets indicate that the pre-processing time remains under 15 seconds per sample with CPU parallel computation. This negligible overhead ensures that the deterministic encoding phase does not become a bottleneck, maintaining the overall efficiency of the PGD-NO pipeline for rapid industrial design iterations. The visualization of the geometry tokens and predicted physics features are shown in Figure~\ref{fig.data_descrip}, and more extracted geometry tokens can be visualized in Appendix~\ref{App.tokens}

\textbf{Baseline}: We used the following model architectures for comparison: PointNet \cite{pointnet}, MeshGraphNet \cite{meshgraphnet}, Graph Neural Operator \cite{GNO}, GALERKIN \cite{galerkin}, Geometry-informed Neural Operator \cite{li2023gino}, General Neural Operator Transformer \cite{GNOT}, Transolver \cite{transolver}, Transolver++ \cite{transolver_plus}, and AB-UPT \cite{AB-UPT}. The details of the hyper-parameter choice of our baseline methods are documented in Appendix~\ref{App.baselines}. To accommodate memory-constrained baselines, the mesh is decomposed into sub-domains for independent inference, then concatenated to reconstruct the global prediction, following the implementations of \cite{transolver_plus}. 

\begin{table*}[t]
  \caption{Comparative analysis of relative error (\%) across 3D industrial benchmarks. Bold values denote the best performance. "-" represents cases where the mesh resolution exceeds the model's memory capacity or the architecture does not support the specific prediction task. In addition to global field errors, we report the prediction accuracy for critical physical measurements (Quantities of Interest) specific to each engineering design problem.}
  \label{tab.main_compare}
  \begin{center}
    \begin{small}
      \begin{sc}
        \begin{tabular}{lcccccccccc}
          \toprule
          \multirow{3}{*}{Model} & \multicolumn{10}{c}{Data Set} \\
           & \multicolumn{2}{c}{Heat Sink} & \multicolumn{2}{c}{JEB} & \multicolumn{2}{c}{DriverNet++} & \multicolumn{2}{c}{Aircraft} & \multicolumn{2}{c}{CFD-VOL} \\ 
           \cmidrule(lr){2-3} \cmidrule(lr){4-5} \cmidrule(lr){6-7} \cmidrule(lr){8-9} \cmidrule(lr){10-11}
           & Field & $\Bar{T}_{\text{chip}}$ & Field & $\sigma_{\text{max}}$ & Field & $C_D$ & Field & $C_L$ & Field & $v_{\text{mid}}$\\
          \midrule
          POINTNET     & 5.21 & 7.54 & 40.1 & 55.4 & 33.4 & 23.7 & 15.2 & 9.15 & 24.5 & 26.2  \\
          MESHGRAPHNET & 3.88 & 4.65 & 55.3 & 62.1 & 38.2 & 28.5 & 10.2 & 3.42 & --    & --     \\
          GNO          & 4.02 & 4.74 & 61.1 & 68.3 & 43.8 & 30.2 & 11.4 & 3.08 & --    & --     \\
          GALERKIN     & 3.52 & 5.15 & 44.2 & 48.9 & 26.4 & 22.1 & 10.1 & 6.53 & --    & --     \\
          GINO         & 2.68 & 4.19 & 38.9 & 44.2 & 28.8 & 24.5 & 11.9 & 4.32 & --    & --     \\
          GNOT         & 3.42 & 5.06 & 41.8 & 47.5 & 21.5 & 22.8 & 8.11 & 3.08 & --    & --     \\
          TRANSOLVER   & 3.20 & 4.89 & 42.6 & 46.3 & 20.1 & 20.3 & 5.23 & 1.41 & --    & --     \\
          AB-UPT & 3.13 & 4.68 & 42.8 & 46.0 & 19.8 & 19.9 & 5.02 & 1.38 & 14.7 & 15.6 \\
          TRANSOLVER++ & 3.09 & 4.57 & 40.3 & 45.2 & 19.2 & 18.5 & 4.98 & 1.35 & 15.8 & 16.7  \\
          \midrule
          PGD-NO-v1 (ours) & \textbf{2.47} & \textbf{3.35} & 39.4 & 43.1 & \textbf{18.2} & \textbf{16.4} & 4.77 & 1.32 & \textbf{12.2} & \textbf{13.2}  \\
          PGD-NO-v2 (ours) & 2.53 & 3.41 & \textbf{37.2} & \textbf{41.7} & 18.9 & 16.9 & 5.37 & 1.39 & 13.9 & 15.3  \\
          PGD-NO-v3 (ours) & --    & --    & --    & --    & 18.5 & 16.3 & \textbf{4.66} & \textbf{1.31} & --    & --     \\
          \bottomrule
        \end{tabular}
      \end{sc}
    \end{small}
  \end{center}
  \vskip -0.1in
\end{table*}

\textbf{Metric}: To quantify the accuracy of our model predictions, we employ the $L^2$ relative error as the primary metric. This provides a holistic assessment of the model's performance over the entire spatial domain. However, global error metrics can sometimes obscure localized physical phenomena that are critical for engineering applications. Therefore, we supplement our analysis by reporting the prediction accuracy for specific critical quantities of interest (QoIs) tailored to each design problem.  More details of the metric are clarified in Appendix~\ref{App.metric}.

\subsection{Main Results}

The experimental results, summarized in Table~\ref{tab.main_compare}, demonstrate that the proposed PGD-NO model consistently outperforms or achieves a similar level of accuracy as established baselines across all five 3D industrial benchmarks. Our architecture achieves the lowest relative errors in both volumetric field predictions and critical global scalar metrics. The performance improvement of PGD-NO over the second-best baseline is most significant on the heat sink dataset. This indicates that when the algorithm can perfectly decompose the geometry into representative tokens, the model is able to learn the underlying partial differential equations (PDEs) much more effectively. Our results on the JEB dataset also indicate that the algorithm is capable of handling distributionally diverse data. However, the improvement over other models in this case is more limited, which suggests a performance plateau when working with such irregular geometries.

% Comparing PGD-NO-v1 and PGD-NO-v2 reveals the effects of increasing model complexity by applying an attention layer to the extracted tokens before mapping the features back to the nodal points. This architectural modification is only helpful for the JEB dataset, where geometry tokens cannot be captured with high precision. Consequently, we consider this design useful primarily when the geometry is not split sufficiently well. In most other cases, this extra complexity may introduce noise into the geometry token representations and decrease overall model performance. 

% Using the third geometry token layer design, we map features back to the points in a hierarchical manner without utilizing long-range dependencies between nodes and non-local tokens. This approach allows PGD-NO-v3 to achieve the best performance on the Aircraft dataset, highlighting that a simpler architecture can be more effective for smooth surface predictions. Finally, our model performs exceptionally well on the CFD-VOL dataset even when a large number of nodes is considered. This demonstrates that utilizing geometry tokens from the surface is sufficient to represent the latent representation of CFD simulation accurately, allowing the model to learn the fluid field data effectively based on these tokens. The error map comparison between our model and Transolver++ are provided in Figure~\ref{fig.error_plot}, which is further calrified in Appendix~\ref{app.err_map}.

Comparing PGD-NO-v1 and v2 reveals that adding an attention layer to geometry tokens primarily benefits datasets like JEB, where complex geometries are harder to partition precisely; otherwise, the extra complexity may introduce noise and degrade performance. Conversely, PGD-NO-v3 hierarchical mapping—which omits long-range dependencies—yields the best results on the Aircraft dataset, suggesting long-range feature relationship are less effective than local relationship for the simple and smooth surfaces. Finally, PGD-NO strong performance on the CFD-VOL dataset demonstrates that surface-derived geometry tokens sufficiently capture the latent representations required for high-fidelity fluid simulations. More results are documented in Appendix~\ref{App.more_results}, including the computational and memory efficiency of different models and comparison with grid-pooling tokens. The error map comparison between our model and Transolver++ are provided in Figure~\ref{fig.error_plot}, which is further clarified in Appendix~\ref{app.err_map}.

\begin{figure*}[t]
  \begin{center}
    \centerline{\includegraphics[width=2.0\columnwidth]{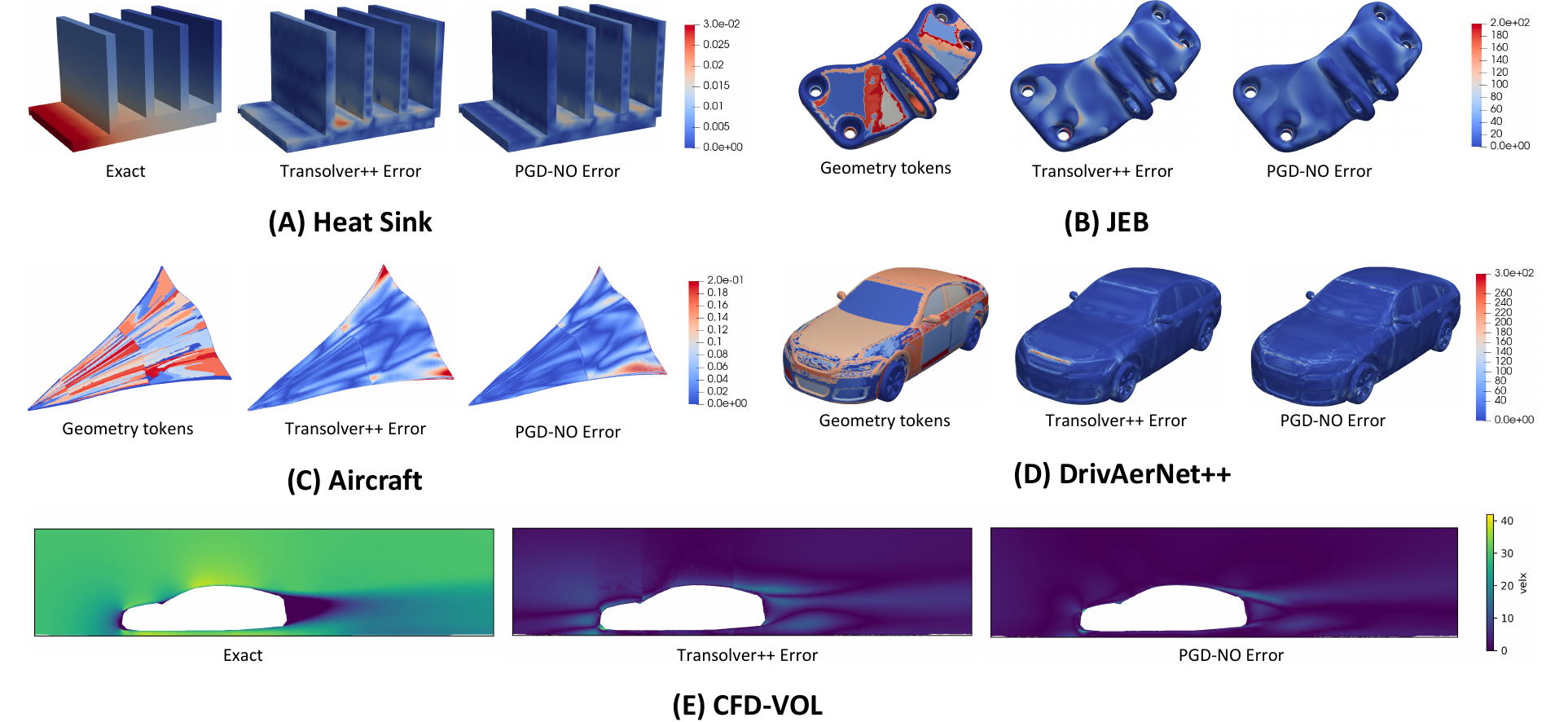}}
    \caption{Error map comparison between Transolver++ and PGD-NO. Geometric references on the left provide spatial context for the residuals. For the Heat Sink and CFD-VOL datasets, we visualize the direct physical fields due to their specific structural or external domain requirements. For all other benchmarks, error maps are overlaid onto the extracted geometry tokens to demonstrate localized model accuracy across distinct structural regions.}
    \label{fig.error_plot}
  \end{center}
\end{figure*}

\begin{figure*}[t]
  \begin{center}
    \centerline{\includegraphics[width=1.8\columnwidth]{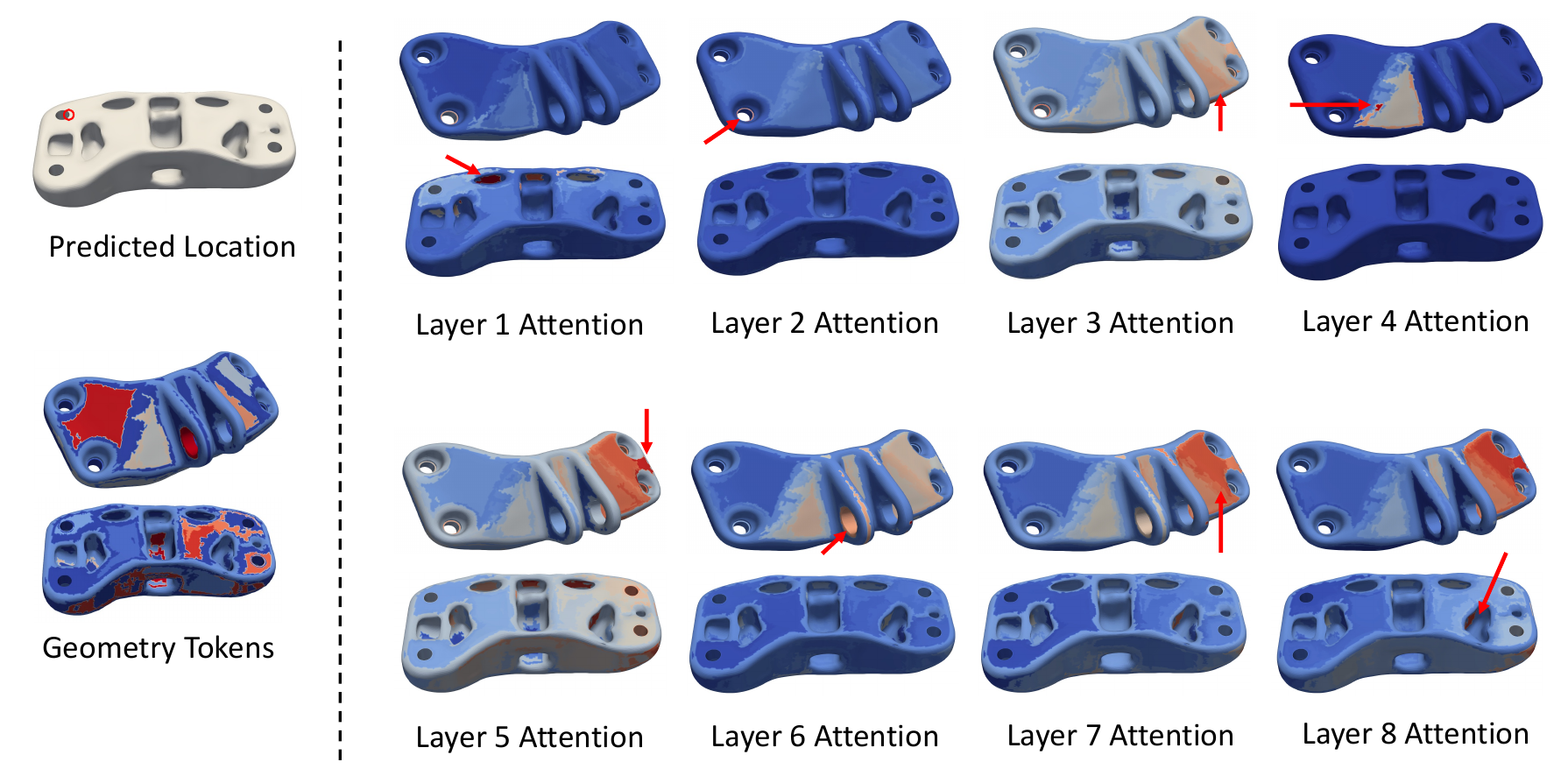}}
    \caption{Interpretability analysis via multi-layer nodal importance mapping. (Left) Geometry reference showing the pre-computed tokens and the specific query location (red circle). (Right) Corresponding nodal importance maps of different geometry token layers, with red arrows highlighting the regions of peak attention.}
    \label{fig.attn_score}
  \end{center}
\end{figure*}

\subsection{Larger Meshes}

To evaluate the scalability of our proposed model on large-scale meshes, we increased the resolution of the CFD-VOL dataset using random spatial sampling and 3D interpolation. As shown in Table~\ref{tab.scaling_results}, we then compared the performance of Transolver++ and PGD-NO-v1 on this up-sampled data.

\begin{table}[h]
\centering
\caption{Comparative relative error (\%) on CFD-VOL across varying mesh resolutions and spatial sampling ratios.}
\label{tab.scaling_results}
\small
\begin{tabular}{ccccccc}
\toprule
Vol. points size & \multicolumn{3}{c}{12M} & \multicolumn{3}{c}{60M} \\ \cmidrule(lr){2-4} \cmidrule(lr){5-7}
Sampling ratio & 0.01 & 0.1 & 1.0 & 0.01 & 0.1 & 1.0 \\ \midrule
Transolver++ & 20.8 & 17.3 & 15.8 & 22.3 & 18.1 & - \\
PGD-NO-v1 & 13.3 & 12.9 & 12.2 & 13.0 & 12.7 & 12.5 \\ \bottomrule
\end{tabular}
\end{table}

While Transolver++ fails to process the full 60M mesh due to memory constraints, PGD-NO-v1 maintains high accuracy even at this scale. To address the challenges of huge dense point clouds, we also utilize a spatial random sampling strategy during training. By computing the loss gradient on a subset of points rather than the entire mesh, we significantly accelerate the training process and reduce peak memory requirements. Our results demonstrate that, Compared with Transolver++,  this stochastic sampling approach does not significantly degrade the gradient quality of our model, as the model continues to deliver high prediction accuracy. Notably, as the total point cloud size increases, the performance gap caused by random sampling is further reduced.

\subsection{Ablation Studies}

Ablation studies on the CFD-VOL dataset (Table~\ref{tab.layers_ablation}) compare the depth sensitivity of Transolver++ and PGD-NO-v1. While Transolver++ shows low sensitivity to increasing layers—likely because its adaptive physics tokens capture complex geometry tokens efficiently in shallower architectures. In contrast, the performance of our model decreases significantly if the number of Geometry Token Layers is insufficient. However, as the number of layers increases, our model demonstrates a superior ability to capture complex geometric features, leading to a consistent and significant enhancement in prediction accuracy. This suggests that the iterative refinement of geometry tokens is vital for improving our model performance.

% We conducted additional ablation studies on the CFD-VOL dataset to evaluate model performance on large-scale point cloud data. While our architecture shares a similar iterative structure with Transolver++ by utilizing a physics attention layer, we specifically employ Geometry Token Layers to capture the underlying physics. A key research question is whether increasing model depth through more layers can provide additional representational capacity. As shown in Table~\ref{tab.layers_ablation}, Transolver++ demonstrates relatively low sensitivity to the number of layers, suggesting the model captures necessary features with a shallower architecture. We attribute this behavior to the adaptive nature of physics token learning, which allows it to represent complex interactions efficiently with fewer iterations. 

% In contrast, the performance of our model decreases significantly if the number of Geometry Token Layers is insufficient. However, as the number of layers increases, our model demonstrates a superior ability to capture complex geometric features, leading to a consistent and significant enhancement in prediction accuracy. This suggests that the iterative refinement of geometry tokens is vital for scaling performance on high-resolution volumetric data.

\begin{table}[h]
\centering
\caption{Comparative $L_2$ relative error (\%) on CFD-VOL across varying model depths.}
\label{tab.layers_ablation}
\small
\begin{tabular}{lcccc}
\toprule
Number of Layers & 2 & 4 & 8 & 12 \\ \midrule
Transolver++ & 17.8 & 16.5 & 15.8 & 15.1 \\
PGD-NO-v1 & 17.9 & 13.5 & 12.2 & 11.1 \\ \bottomrule
\end{tabular}
\end{table}

We conduct an additional ablation study of number of tokens using the JEB dataset, as it contains the most complex and organic geometries in our benchmark. As shown in Table~\ref{tab:num_token_ablation}, while performance degrades significantly with an insufficient number of tokens, marginal gains diminish as the token count scales up, which indicates that our proposed method can provide stable performance as sufficient geometry tokens are extracted.

\begin{table}[h]
\centering
\caption{Comparative $L_2$ relative error (\%) on JEB across varying number of extracted geometry tokens.}
\label{tab:num_token_ablation}
\begin{tabular}{lcccc}
\toprule
Number of Tokens & 32 & 64 & 128 & 256 \\ 
\midrule
PGD-NO (Ours) & 65.4 & 41.5 & 37.2 & 35.4 \\ 
\bottomrule
\end{tabular}
\end{table}

\subsection{Model Explanation}

% To compute nodal importance scores, our model utilizes a multi-layer cross-attention mechanism within the iterative Geometric Token Layers. We first extract the attention weights from the cross-attention module to determine the contribution of each token to a specific query point. These token-level scores are then uniformly distributed among the constituent nodes of each token. 

Our model produces granular, node-wise importance maps by redistributing cross-attention weights from the Geometric Token Layers to their constituent nodes—a diagnostic capability absent in existing baseline models. In Figure~\ref{fig.attn_score}, we visualize the nodal attention score distribution relative to the predicted location of maximum stress, with the peak attention region highlighted by a red arrow. The visualization reveals several key behaviors of the model. First, the model focuses on localized tokens near the predicted region before attending to larger, distant tokens. Generally, nodes on the left side of the bracket, which are closer to the region of interest, receive higher attention scores than those on the right. Notably, the model prioritizes the constraint region in Layer 2 and the loading region in Layer 6. This aligns with physical intuition, as these areas define the boundary conditions that govern the stress distribution. Each layer targets distinct spatial locations, indicating that the model successfully integrates multi-scale geometric details to inform its final prediction. These high-attention regions can ultimately serve as a design heuristic, guiding engineers toward critical structural areas that require optimization.

% Our model generates granular node-wise importance maps by redistributing cross-attention weights from the Geometric Token Layers to their constituent nodes. As shown in Figure~\ref{fig.attn_score}, the model prioritizes localized tokens near the region of interest before attending to larger, distant structures. Specifically, Layer 2 focuses on constraint regions while Layer 6 targets loading zones, aligning with the physical boundary conditions governing stress distribution. This multi-scale attention serves as a diagnostic heuristic, guiding structural optimization by identifying critical geometric features across successive layers.

\section{Conclusion}

We introduce the Pre-computed Geometry Decomposition Neural Operator (PGD-NO), a paradigm that mitigates GPU memory bottlenecks in high-fidelity PDE simulations by shifting geometric encoding to a deterministic, pre-computed step. By extracting "geometry tokens" that effectively represent latent physical features, our model achieves linear scalability and handles massive meshes—up to 100 million nodes without inter-GPU communication overhead. Our results across five diverse industrial benchmarks demonstrate that PGD-NO can achieve similar accuracy as state-of-the-art baselines and provides enhanced interpretability.

\textbf{Limitations}: While PGD-NO matches state-of-the-art accuracy and offers enhanced interpretability via attention-based importance scores, combining our structural decomposition with local learnable refinements can be further investigated to better represent irregular, smooth topologies while maintaining the efficiency of decoupled volume-surface querying. Additionally, integrating recent advanced point cloud sampling techniques into our framework could further boost computational efficiency. Finally, because PGD-NO relies on precomputed geometry tokens, it exhibits robust invariance to underlying geometry representation. This decoupled nature makes our approach a promising candidate for building scientific foundation models capable of handling diverse data sources with varying geometric design principles and meshing formats—an avenue we intend to investigate thoroughly in the future.

\section*{Impact Statement}

This paper presents work aimed at advancing the efficiency of neural operators for 3D physics deep learning. Our method enables large-scale physical simulations on complex industrial geometries that were previously computationally prohibitive, offering a generalizable surrogate modeling framework applicable to most current industry-scale modeling problems. Beyond accelerating industrial design and manufacturing, our focus on 3D geometry learning provides valuable insights for research into the development of geometry generation models and other frameworks centered on 3D geometries.

\newpage

\bibliography{example_paper}

@inproceedings{GNOT,
  title={Gnot: A general neural operator transformer for operator learning},
  author={Hao, Zhongkai and Wang, Zhengyi and Su, Hang and Ying, Chengyang and Dong, Yinpeng and Liu, Songming and Cheng, Ze and Song, Jian and Zhu, Jun},
  booktitle={International Conference on Machine Learning},
  pages={12556--12569},
  year={2023},
  organization={PMLR}
}

@article{fem,
  title={Finite element method: an overview},
  author={Jagota, Vishal and Sethi, Aman Preet Singh and Kumar, Khushmeet},
  journal={Walailak Journal of Science and Technology (WJST)},
  volume={10},
  number={1},
  pages={1--8},
  year={2013}
}

@article{S-DON,
  title={Sequential deep operator networks (s-deeponet) for predicting full-field solutions under time-dependent loads},
  author={He, Junyan and Kushwaha, Shashank and Park, Jaewan and Koric, Seid and Abueidda, Diab and Jasiuk, Iwona},
  journal={Engineering Applications of Artificial Intelligence},
  volume={127},
  pages={107258},
  year={2024},
  publisher={Elsevier}
}

@article{DCON,
  title={Physics-informed discretization-independent deep compositional operator network},
  author={Zhong, Weiheng and Meidani, Hadi},
  journal={Computer Methods in Applied Mechanics and Engineering},
  volume={431},
  pages={117274},
  year={2024},
  publisher={Elsevier}
}

@inproceedings{GNO,
  title={Neural operator: Graph kernel network for partial differential equations},
  author={Anandkumar, Anima and Azizzadenesheli, Kamyar and Bhattacharya, Kaushik and Kovachki, Nikola and Li, Zongyi and Liu, Burigede and Stuart, Andrew},
  booktitle={ICLR 2020 workshop on integration of deep neural models and differential equations},
  year={2020}
}

@inproceedings{drivAer,
  title={Drivaernet: A parametric car dataset for data-driven aerodynamic design and graph-based drag prediction},
  author={Elrefaie, Mohamed and Dai, Angela and Ahmed, Faez},
  booktitle={International Design Engineering Technical Conferences and Computers and Information in Engineering Conference},
  volume={88360},
  pages={V03AT03A019},
  year={2024},
  organization={American Society of Mechanical Engineers}
}

@article{drivAer++,
  title={Drivaernet++: A large-scale multimodal car dataset with computational fluid dynamics simulations and deep learning benchmarks},
  author={Elrefaie, Mohamed and Morar, Florin and Dai, Angela and Ahmed, Faez},
  journal={Advances in Neural Information Processing Systems},
  volume={37},
  pages={499--536},
  year={2024}
}

@article{deepJEB,
  title={Deepjeb: 3d deep learning-based synthetic jet engine bracket dataset},
  author={Hong, Seongjun and Kwon, Yongmin and Shin, Dongju and Park, Jangseop and Kang, Namwoo},
  journal={Journal of Mechanical Design},
  volume={147},
  number={4},
  year={2025},
  publisher={American Society of Mechanical Engineers Digital Collection}
}

@article{pointnet,
  title={Physics-informed PointNet: A deep learning solver for steady-state incompressible flows and thermal fields on multiple sets of irregular geometries},
  author={Kashefi, Ali and Mukerji, Tapan},
  journal={Journal of Computational Physics},
  volume={468},
  pages={111510},
  year={2022},
  publisher={Elsevier}
}

@article{EAGNO,
  title={Mesh-based GNN surrogates for time-independent PDEs},
  author={Gladstone, Rini Jasmine and Rahmani, Helia and Suryakumar, Vishvas and Meidani, Hadi and D’Elia, Marta and Zareei, Ahmad},
  journal={Scientific reports},
  volume={14},
  number={1},
  pages={3394},
  year={2024},
  publisher={Nature Publishing Group UK London}
}

@article{GANO,
  title={Physics-informed geometry-aware neural operator},
  author={Zhong, Weiheng and Meidani, Hadi},
  journal={Computer Methods in Applied Mechanics and Engineering},
  volume={434},
  pages={117540},
  year={2025},
  publisher={Elsevier}
}

@article{Neural_field,
  title={Operator learning with neural fields: Tackling pdes on general geometries},
  author={Serrano, Louis and Le Boudec, Lise and Kassa{\"\i} Koupa{\"\i}, Armand and Wang, Thomas X and Yin, Yuan and Vittaut, Jean-No{\"e}l and Gallinari, Patrick},
  journal={Advances in Neural Information Processing Systems},
  volume={36},
  pages={70581--70611},
  year={2023}
}

@incollection{fvm,
  title={The finite volume method},
  author={Moukalled, Fadl and Mangani, Luca and Darwish, Marwan},
  booktitle={The finite volume method in computational fluid dynamics: An advanced introduction with OpenFOAM{\textregistered} and Matlab},
  pages={103--135},
  year={2015},
  publisher={Springer}
}

@inproceedings{nvlink,
  title={9.3 NVLink-C2C: A coherent off package chip-to-chip interconnect with 40Gbps/pin single-ended signaling},
  author={Wei, Ying and Huang, Yi Chieh and Tang, Haiming and Sankaran, Nithya and Chadha, Ish and Dai, Dai and Oluwole, Olakanmi and Balan, Vishnu and Lee, Edward},
  booktitle={2023 IEEE International Solid-State Circuits Conference (ISSCC)},
  pages={160--162},
  year={2023},
  organization={IEEE}
}

@article{galerkin,
  title={Choose a transformer: Fourier or galerkin},
  author={Cao, Shuhao},
  journal={Advances in neural information processing systems},
  volume={34},
  pages={24924--24940},
  year={2021}
}

@article{DON,
  title={Learning nonlinear operators via DeepONet based on the universal approximation theorem of operators},
  author={Lu, Lu and Jin, Pengzhan and Pang, Guofei and Zhang, Zhongqiang and Karniadakis, George Em},
  journal={Nature machine intelligence},
  volume={3},
  number={3},
  pages={218--229},
  year={2021},
  publisher={Nature Publishing Group UK London}
}

@inproceedings{FNO,
  title={Burigede liu, Kaushik Bhattacharya, Andrew Stuart, and Anima Anandkumar. Fourier neural operator for parametric partial differential equations},
  author={Li, Zongyi and Kovachki, Nikola Borislavov and Azizzadenesheli, Kamyar},
  booktitle={International Conference on Learning Representations},
  volume={2},
  number={3},
  pages={8},
  year={2021}
}

@article{li2020multipole,
  title={Multipole graph neural operator for parametric partial differential equations},
  author={Li, Zongyi and Kovachki, Nikola and Azizzadenesheli, Kamyar and Liu, Burigede and Stuart, Andrew and Bhattacharya, Kaushik and Anandkumar, Anima},
  journal={Advances in Neural Information Processing Systems},
  volume={33},
  pages={6755--6766},
  year={2020}
}

@article{lno2024,
  title={Latent neural operator for solving forward and inverse pde problems},
  author={Wang, Tian and Wang, Chuang},
  journal={Advances in Neural Information Processing Systems},
  volume={37},
  pages={33085--33107},
  year={2024}
}

@article{li2023gino,
  title={Geometry-informed neural operator for large-scale 3d pdes},
  author={Li, Zongyi and Kovachki, Nikola and Choy, Chris and Li, Boyi and Kossaifi, Jean and Otta, Shourya and Nabian, Mohammad Amin and Stadler, Maximilian and Hundt, Christian and Azizzadenesheli, Kamyar and others},
  journal={Advances in Neural Information Processing Systems},
  volume={36},
  pages={35836--35854},
  year={2023}
}

@article{eliasof2021pde,
  title={Pde-gcn: Novel architectures for graph neural networks motivated by partial differential equations},
  author={Eliasof, Moshe and Haber, Eldad and Treister, Eran},
  journal={Advances in neural information processing systems},
  volume={34},
  pages={3836--3849},
  year={2021}
}

@article{AB-UPT,
title={{AB}-{UPT}: Scaling Neural {CFD} Surrogates for High- Fidelity Automotive Aerodynamics Simulations via Anchored- Branched Universal Physics Transformers},
author={Benedikt Alkin and Maurits Bleeker and Richard Kurle and Tobias Kronlachner and Reinhard Sonnleitner and Matthias Dorfer and Johannes Brandstetter},
journal={Transactions on Machine Learning Research},
issn={2835-8856},
year={2025},
url={https://openreview.net/forum?id=nwQ8nitlTZ},
note={}
}

@article{velasco2024gtnn,
  title={Graph neural networks and non-commuting operators},
  author={Velasco, Mauricio and O'Hare, Kaiying and Rychtenberg, Bernardo and Villar, Soledad},
  journal={Advances in neural information processing systems},
  volume={37},
  pages={95662--95691},
  year={2024}
}

@article{lu2021,
  title={Learning nonlinear operators via DeepONet based on the universal approximation theorem of operators},
  author={Lu, Lu and Jin, Pengzhan and Pang, Guofei and Zhang, Zhongqiang and Karniadakis, George Em},
  journal={Nature machine intelligence},
  volume={3},
  number={3},
  pages={218--229},
  year={2021},
  publisher={Nature Publishing Group UK London}
}

@article{pivae,
  title={Pi-vae: Physics-informed variational auto-encoder for stochastic differential equations},
  author={Zhong, Weiheng and Meidani, Hadi},
  journal={Computer Methods in Applied Mechanics and Engineering},
  volume={403},
  pages={115664},
  year={2023},
  publisher={Elsevier}
}

@article{Mandl2024,
  title={Separable physics-informed DeepONet: Breaking the curse of dimensionality in physics-informed machine learning},
  author={Mandl, Luis and Goswami, Somdatta and Lambers, Lena and Ricken, Tim},
  journal={Computer Methods in Applied Mechanics and Engineering},
  volume={434},
  pages={117586},
  year={2025},
  publisher={Elsevier}
}

@article{lone2026alpha,
  title={$\alpha$-VI DeepONet: A prior-robust variational Bayesian approach for enhancing DeepONets with uncertainty quantification},
  author={Lone, Soban Nasir and De, Subhayan and Nayek, Rajdip},
  journal={Computer Methods in Applied Mechanics and Engineering},
  volume={449},
  pages={118552},
  year={2026},
  publisher={Elsevier}
}

@article{Zhang2025,
  title={Deeponet as a multi-operator extrapolation model: Distributed pretraining with physics-informed fine-tuning},
  author={Zhang, Zecheng and Moya, Christian and Lu, Lu and Lin, Guang and Schaeffer, Hayden},
  journal={Journal of Computational Physics},
  pages={114537},
  year={2025},
  publisher={Elsevier}
}

@article{derivative_don,
  title={Derivative-enhanced deep operator network},
  author={Qiu, Yuan and Bridges, Nolan and Chen, Peng},
  journal={Advances in Neural Information Processing Systems},
  volume={37},
  pages={20945--20981},
  year={2024}
}

@article{Park2025,
  title={Point-DeepONet: Predicting Nonlinear Fields on Non-Parametric Geometries under Variable Load Conditions},
  author={Park, Jangseop and Kang, Namwoo},
  journal={Neural Networks},
  pages={108560},
  year={2026},
  publisher={Elsevier}
}

@inproceedings{adam,
  title={A proof of local convergence for the Adam optimizer},
  author={Bock, Sebastian and Wei{\ss}, Martin},
  booktitle={2019 international joint conference on neural networks (IJCNN)},
  pages={1--8},
  year={2019},
  organization={IEEE}
}

@article{hashgrid,
  title={Hashgrid: an optimized architecture for accelerating graph computing on fpgas},
  author={Sahebi, Amin and Procaccini, Marco and Giorgi, Roberto},
  journal={Future Generation Computer Systems},
  volume={162},
  pages={107497},
  year={2025},
  publisher={Elsevier}
}

@article{D-FNO2025,
  title={D-FNO: A decomposed Fourier neural operator for large-scale parametric partial differential equations},
  author={Li, Kangjie and Ye, Wenjing},
  journal={Computer Methods in Applied Mechanics and Engineering},
  volume={436},
  pages={117732},
  year={2025},
  publisher={Elsevier}
}

@article{DAFNO2023,
  title={Domain agnostic fourier neural operators},
  author={Liu, Ning and Jafarzadeh, Siavash and Yu, Yue},
  journal={Advances in Neural Information Processing Systems},
  volume={36},
  pages={47438--47450},
  year={2023}
}

@article{PINO2021,
  title={Physics-informed neural operator for learning partial differential equations},
  author={Li, Zongyi and Zheng, Hongkai and Kovachki, Nikola and Jin, David and Chen, Haoxuan and Liu, Burigede and Azizzadenesheli, Kamyar and Anandkumar, Anima},
  journal={ACM/JMS Journal of Data Science},
  volume={1},
  number={3},
  pages={1--27},
  year={2024},
  publisher={ACM New York, NY}
}

@article{BenchmarksFNO2021,
  title={A comprehensive and fair comparison of two neural operators (with practical extensions) based on fair data},
  author={Lu, Lu and Meng, Xuhui and Cai, Shengze and Mao, Zhiping and Goswami, Somdatta and Zhang, Zhongqiang and Karniadakis, George Em},
  journal={Computer Methods in Applied Mechanics and Engineering},
  volume={393},
  pages={114778},
  year={2022},
  publisher={Elsevier}
}

@article{li2023scalable,
  title={Scalable transformer for pde surrogate modeling},
  author={Li, Zijie and Shu, Dule and Barati Farimani, Amir},
  journal={Advances in Neural Information Processing Systems},
  volume={36},
  pages={28010--28039},
  year={2023}
}

@InProceedings{transolver,
  title = 	 {Transolver: A Fast Transformer Solver for {PDE}s on General Geometries},
  author =       {Wu, Haixu and Luo, Huakun and Wang, Haowen and Wang, Jianmin and Long, Mingsheng},
  booktitle = 	 {Proceedings of the 41st International Conference on Machine Learning},
  pages = 	 {53681--53705},
  year = 	 {2024},
  editor = 	 {Salakhutdinov, Ruslan and Kolter, Zico and Heller, Katherine and Weller, Adrian and Oliver, Nuria and Scarlett, Jonathan and Berkenkamp, Felix},
  volume = 	 {235},
  series = 	 {Proceedings of Machine Learning Research},
  month = 	 {21--27 Jul},
  publisher =    {PMLR},
  url = 	 {https://proceedings.mlr.press/v235/wu24r.html}
}

@article{alkin2024universal,
  title={Universal physics transformers: A framework for efficiently scaling neural operators},
  author={Alkin, Benedikt and F{\"u}rst, Andreas and Schmid, Simon and Gruber, Lukas and Holzleitner, Markus and Brandstetter, Johannes},
  journal={Advances in Neural Information Processing Systems},
  volume={37},
  pages={25152--25194},
  year={2024}
}

@article{zheng2024alias,
  title={Alias-Free Mamba Neural Operator},
  author={Zheng, Jianwei and Li, Wei and Xu, Ni and Zhu, Junwei and Zhang, Xiaoqin},
  journal={Advances in Neural Information Processing Systems},
  volume={37},
  pages={52962--52995},
  year={2024}
}

@article{yu2024nonlocal,
  title={Nonlocal attention operator: Materializing hidden knowledge towards interpretable physics discovery},
  author={Yu, Yue and Liu, Ning and Lu, Fei and Gao, Tian and Jafarzadeh, Siavash and Silling, Stewart A},
  journal={Advances in Neural Information Processing Systems},
  volume={37},
  pages={113797--113822},
  year={2024}
}

@article{morrison2024gfn,
  title={GFN: A graph feedforward network for resolution-invariant reduced operator learning in multifidelity applications},
  author={Morrison, Ois{\'\i}n M and Pichi, Federico and Hesthaven, Jan S},
  journal={Computer Methods in Applied Mechanics and Engineering},
  volume={432},
  pages={117458},
  year={2024},
  publisher={Elsevier}
}

@inproceedings{lee2024inducing,
  title={Inducing point operator transformer: A flexible and scalable architecture for solving pdes},
  author={Lee, Seungjun and Oh, Taeil},
  booktitle={Proceedings of the AAAI Conference on Artificial Intelligence},
  volume={38},
  number={1},
  pages={153--161},
  year={2024}
}

@data{heatsink-dataset,
author = {Zhong, Weiheng},
publisher = {Harvard Dataverse},
title = {{Heatsink}},
year = {2025},
version = {V1},
doi = {10.7910/DVN/OZQMD4},
url = {https://doi.org/10.7910/DVN/OZQMD4}
}

@article{fem_sensitivity,
  title={Recent progress in nonlinear FEM-based sensitivity analysis},
  author={Hisada, Toshiaki},
  journal={JSME international journal. Ser. A, Mechanics and material engineering},
  volume={38},
  number={3},
  pages={301--310},
  year={1995},
  publisher={The Japan Society of Mechanical Engineers}
}

@InProceedings{transolver_plus,
  title = 	 {Transolver++: An Accurate Neural Solver for {PDE}s on Million-Scale Geometries},
  author =       {Luo, Huakun and Wu, Haixu and Zhou, Hang and Xing, Lanxiang and Di, Yichen and Wang, Jianmin and Long, Mingsheng},
  booktitle = 	 {Proceedings of the 42nd International Conference on Machine Learning},
  pages = 	 {41432--41449},
  year = 	 {2025},
  editor = 	 {Singh, Aarti and Fazel, Maryam and Hsu, Daniel and Lacoste-Julien, Simon and Berkenkamp, Felix and Maharaj, Tegan and Wagstaff, Kiri and Zhu, Jerry},
  volume = 	 {267},
  series = 	 {Proceedings of Machine Learning Research},
  month = 	 {13--19 Jul},
  publisher =    {PMLR},
  url = 	 {https://proceedings.mlr.press/v267/luo25o.html}
}

@article{tradition_geo_decomp,
  title={Decomposition of geometric constraint systems: a survey},
  author={Jermann, Christophe and Trombettoni, Gilles and Neveu, Bertrand and Mathis, Pascal},
  journal={International Journal of Computational Geometry \& Applications},
  volume={16},
  number={05n06},
  pages={379--414},
  year={2006},
  publisher={World Scientific}
}

@inproceedings{mesh_based_decomp,
  title={A simple and efficient approach for 3D mesh approximate convex decomposition},
  author={Mamou, Khaled and Ghorbel, Faouzi},
  booktitle={2009 16th IEEE international conference on image processing (ICIP)},
  pages={3501--3504},
  year={2009},
  organization={IEEE}
}

@inproceedings{laplace_geo_decom,
  title={Laplace-beltrami eigenfunctions towards an algorithm that" understands" geometry},
  author={L{\'e}vy, Bruno},
  booktitle={IEEE International Conference on Shape Modeling and Applications 2006 (SMI'06)},
  pages={13--13},
  year={2006},
  organization={IEEE}
}

@article{convex_geo_decomp,
  title={Approximate convex decomposition of polyhedra and its applications},
  author={Lien, Jyh-Ming and Amato, Nancy M},
  journal={Computer Aided Geometric Design},
  volume={25},
  number={7},
  pages={503--522},
  year={2008},
  publisher={Elsevier}
}

@article{shatter,
  title={From segmentation to shattering: Structural transitions in the breakup of brittle rings},
  author={Szuszik, Csan{\'a}d and Szatm{\'a}ri, Roland and P{\'a}l, Gerg{\H{o}} and Kun, Ferenc},
  journal={Chaos, Solitons \& Fractals},
  volume={202},
  pages={117573},
  year={2026},
  publisher={Elsevier}
}

@inproceedings{
  meshgraphnet,
  title={Learning Mesh-Based Simulation with Graph Networks},
  author={Tobias Pfaff and Meire Fortunato and Alvaro Sanchez-Gonzalez and Peter W. Battaglia},
  booktitle={International Conference on Learning Representations (ICLR)},
  year={2021}
}
\bibliographystyle{icml2026}

%%%%%%%%%%%%%%%%%%%%%%%%%%%%%%%%%%%%%%%%%%%%%%%%%%%%%%%%%%%%%%%%%%%%%%%%%%%%%%%
%%%%%%%%%%%%%%%%%%%%%%%%%%%%%%%%%%%%%%%%%%%%%%%%%%%%%%%%%%%%%%%%%%%%%%%%%%%%%%%
% APPENDIX
%%%%%%%%%%%%%%%%%%%%%%%%%%%%%%%%%%%%%%%%%%%%%%%%%%%%%%%%%%%%%%%%%%%%%%%%%%%%%%%
%%%%%%%%%%%%%%%%%%%%%%%%%%%%%%%%%%%%%%%%%%%%%%%%%%%%%%%%%%%%%%%%%%%%%%%%%%%%%%%
\newpage
\appendix
\onecolumn

\section{Model complexity analysis \label{App.complexity_analysis}}

In the Transolver++ framework, each physics-based attention layer requires approximately eight $O(N)$ operations per volume node. Specifically, for every block, the model must perform an \textbf{input projection} of nodal features to the internal head dimension, followed by \textbf{two separate linear mappings} to determine the Gumbel-Softmax temperature and the slice membership weights. The membership assignment itself necessitates a \textbf{Softmax operation} across the entire node set. The core interaction involves \textbf{two Einstein summation steps}: one to aggregate $N$ nodes into a reduced set of geometry tokens, and another to redistribute the processed token features back to the original $N$ coordinates. Finally, the layer concludes with an \textbf{output linear projection} and a subsequent \textbf{MLP pass} that maps the hidden representation back to the nodal feature space. In contrast, our first two proposed attention layer architectures, as illustrated in Figure~\ref{fig.model_arch} (B), optimize this process by requiring only three primary operations: query projection, attention score computation, and the linear operation for final output in a Multi-Head Attention layer. This architectural effectively reduces the computational overhead by more than 50\% while maintaining representative capacity.

Regarding memory complexity, our model shifts the primary bottleneck from the volume nodes to the surface nodes. While Transolver++ operates on the entire point cloud holistically, resulting in a memory footprint dominated by the number of volume nodes ($O(8N_v)$), our framework decouples geometry feature extraction from solution querying. Because volume nodes in our model serve strictly as query points for PDE evaluation and do not participate in surface feature encoding, the volume point cloud can be partitioned and processed independently. In an extreme parallelization scenario, each volume node could theoretically be processed by a discrete unit. Consequently, the memory bottleneck resides within the surface node operations of the Geometry Token Layer. With a memory complexity of $O(3N_s)$, and utilizing hardware such as the NVIDIA GH200, our framework supports efficient parallel training for high-fidelity meshes containing up to 80 million surface nodes.

\section{Dataset description \label{App.data}}

The details of the physics simulation are clarified:

\begin{itemize}
    \item \textbf{Heat Sink (Thermal Management):} In the semiconductor industry, precise thermal management is essential to prevent hardware failure due to overheating. This simulation models steady-state heat dissipation from a CPU through a parameterized geometry. The design is parameterized by $\mathbf{p} \in \mathbb{R}^4$ (fin height, thickness, spacing, and count) mapped to a 10k-node 3D coordinate mesh $\mathcal{X}$. The output is the steady-state temperature field $T(\mathbf{x})$, governed by the Poisson equation for heat conduction:
    \[ -\nabla \cdot (\kappa \nabla T) = Q \]
    where $\kappa$ is the thermal conductivity and $Q$ is the heat source. Boundary conditions specify forced convection $q = h(T - T_{\infty})$ on the lateral surfaces and free convection on the top surface.

    \item \textbf{JEB (Aerospace Structural Engineering):} Reducing component weight while maintaining structural integrity is a core objective in aerospace manufacturing. This dataset utilizes 2,138 freeform 3D bracket designs where the input is a diverse set of 100k-scale meshes $\mathcal{M}$ with fixed bolt-hole constraints. The output is the displacement vector field $\mathbf{u}(\mathbf{x})$ and the stress distribution, derived from the linear elasticity equilibrium equation:
    \[ \nabla \cdot \sigma + \mathbf{f} = 0 \]
    where $\sigma$ is the stress tensor and $\mathbf{f}$ represents the applied vertical loading of 3.5 kN. This setup tests the model's ability to solve for structural stability under prescribed mechanical loads.

    \item \textbf{DrivAer++ (Automotive Aerodynamics):} Minimizing aerodynamic drag is critical for automotive OEMs to meet fuel efficiency and emission standards. The input consists of high-fidelity surface meshes of realistic car designs, including complex features like wheels and underbodies. The output is the surface pressure $p$ and wall shear stress $\tau_w$ fields, which are solutions to the Reynolds-Averaged Navier-Stokes (RANS) equations for turbulent flow:
    \[ \rho (\mathbf{u} \cdot \nabla) \mathbf{u} = -\nabla p + \mu \nabla^2 \mathbf{u} + \nabla \cdot \tau^R \]
    where $\tau^R$ is the Reynolds stress tensor. Boundary conditions include a defined inlet velocity and no-slip conditions on the vehicle surface.

    \item \textbf{Aircraft (Transonic Aerodynamics):} Aircraft design requires evaluating aerodynamic performance across varying flight conditions to ensure stability and efficiency. The input features include a 300,000-point surface mesh alongside flow conditions: Mach number $M$, angle of attack $\alpha$, and sideslip angle $\beta$. The output is a multi-variable state vector including fluid density $\rho$, pressure $p$, and velocity components $(u, v, w)$, governed by the compressible Navier-Stokes equations:
    \[ \frac{\partial \rho}{\partial t} + \nabla \cdot (\rho \mathbf{u}) = 0 \]
    This benchmark captures the complex physics of transonic flow, specifically the formation and movement of shock waves on the airframe surfaces.

    \item \textbf{CFD-VOL (Industrial Fluid Mechanics):} Full volumetric fluid analysis is required for comprehensive engineering tasks such as HVAC design and internal flow modeling. The output of the CFD simulation of the fluid part is the 3D velocity field $\mathbf{u}(\mathbf{x})$ on a 12M-node-scale mesh and pressure field $p(\mathbf{x})$ throughout the entire volume, governed by the incompressible Navier-Stokes and continuity equations:
    \[ \rho (\mathbf{u} \cdot \nabla) \mathbf{u} = -\nabla p + \mu \nabla^2 \mathbf{u} + \mathbf{f}, \quad \nabla \cdot \mathbf{u} = 0 \]
    This setup focuses on the extreme scalability and local conservation required in large-scale industrial fluid simulations.
\end{itemize}

\section{Additional results \label{App.more_results}}

In addition to accuracy comparison, we provide a detailed comparison of training times and memory usage in Table~\ref{tab:efficiency_comparison}. Entries with "-" indicate memory requirements exceeding the a 4-way GH200 capacity (approx. 360GB).

\begin{table*}[htbp]
\centering
\caption{Comparison of Training Times and Memory Usage Across Datasets}
\label{tab:efficiency_comparison}
\small % Reduces font size slightly to fit all columns comfortably
\setlength{\tabcolsep}{5pt} % Adjusts spacing between columns
\begin{tabular}{lcccccccc}
\toprule
\textbf{Dataset} & \textbf{PointNet} & \textbf{GNO} & \textbf{MeshGraphNet} & \textbf{Galerkin} & \textbf{GINO} & \textbf{GNOT} & \textbf{Transolver++} & \textbf{PGD-NO (Ours)} \\ 
\midrule
Heatsink & 800s & 9,000s & 8,450s & 8,660s & 9,200s & 10,800s & 7,800s & 6,200s \\ 
\addlinespace
JEB & 670s & 7,400s & 7,800s & 22,600s & 19,400s & 15,100s & 14,500s & 14,000s \\ 
\addlinespace
Aircraft & 1.6h (15G) & 28.2h (-) & 27.8h (-) & 16.4h (-) & 7.8h (-) & 8.1h (89G) & 6.2h (77G) & 5.4h (42G) \\ 
\addlinespace
Drivernet & 2.1h (10G) & 37.5h (-) & 37.3h (-) & 11.6h (89G) & 12.9h (-) & 10.2h (75G) & 9.5h (62G) & 8.8h (35GB) \\ 
\addlinespace
CFD-VOL & 13.2h (30G) & -- & -- & -- & -- & -- & 24.3h (345G) & 22h (120G) \\ 
\bottomrule
\end{tabular}
\end{table*}

To demonstrate the importance of our geometry decomposition for model performance, we compare it against a baseline where geometry latent tokens are instead generated via mean-pooling of the point cloud features. Specifically, this baseline partitions the point cloud into uniform grid cells, and compute the average of the point features inside each cell and treat it as one geometry token. Then we use the same decoder structure as our PGD-NO-v1 to generate predictions. As shown in the Table~\ref{tab:grid_vs_surface_tokens}, this grid-pooling approach consistently underperforms across all datasets in terms of $L_2$ relative error when compared to our proposed geometry decomposition (Surface token). 

\begin{table}[htbp]
\centering
\caption{Comparison between surface tokens and grid-pooling tokens}
\label{tab:grid_vs_surface_tokens}
\begin{tabular}{lccccc}
\toprule
\textbf{Method} & \textbf{Heatsink} & \textbf{JEB} & \textbf{Aircraft} & \textbf{Drivernet} & \textbf{CFD-VOL} \\ 
\midrule
Grid- token & 9.43 & 62.5 & 35.4 & 10.5 & 36.7 \\ 
Surface token (Ours) & 2.47 & 37.2 & 18.2 & 4.66 & 12.2 \\ 
\bottomrule
\end{tabular}
\end{table}

\newpage

\section{Extracted geometry token visualizations \label{App.tokens}}

\begin{figure}[H]
  \begin{center}
    \centerline{\includegraphics[width=0.55\columnwidth]{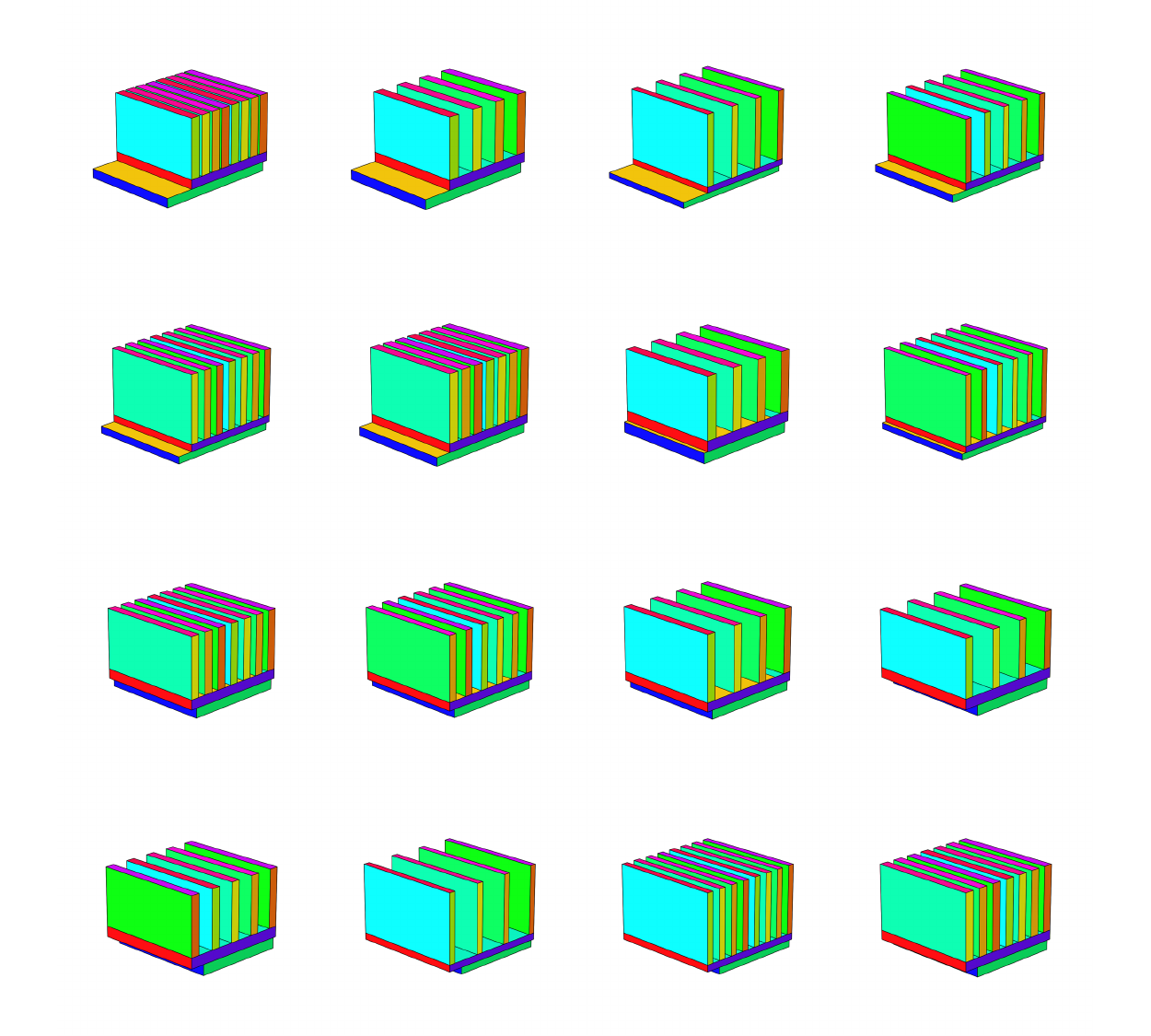}}
    \caption{The extracted tokens of different samples in Heat Sink dataset are presented.}
    \label{fig.heatsink_tokens}
  \end{center}
\end{figure}

\begin{figure}[H]
  \begin{center}
    \centerline{\includegraphics[width=0.55\columnwidth]{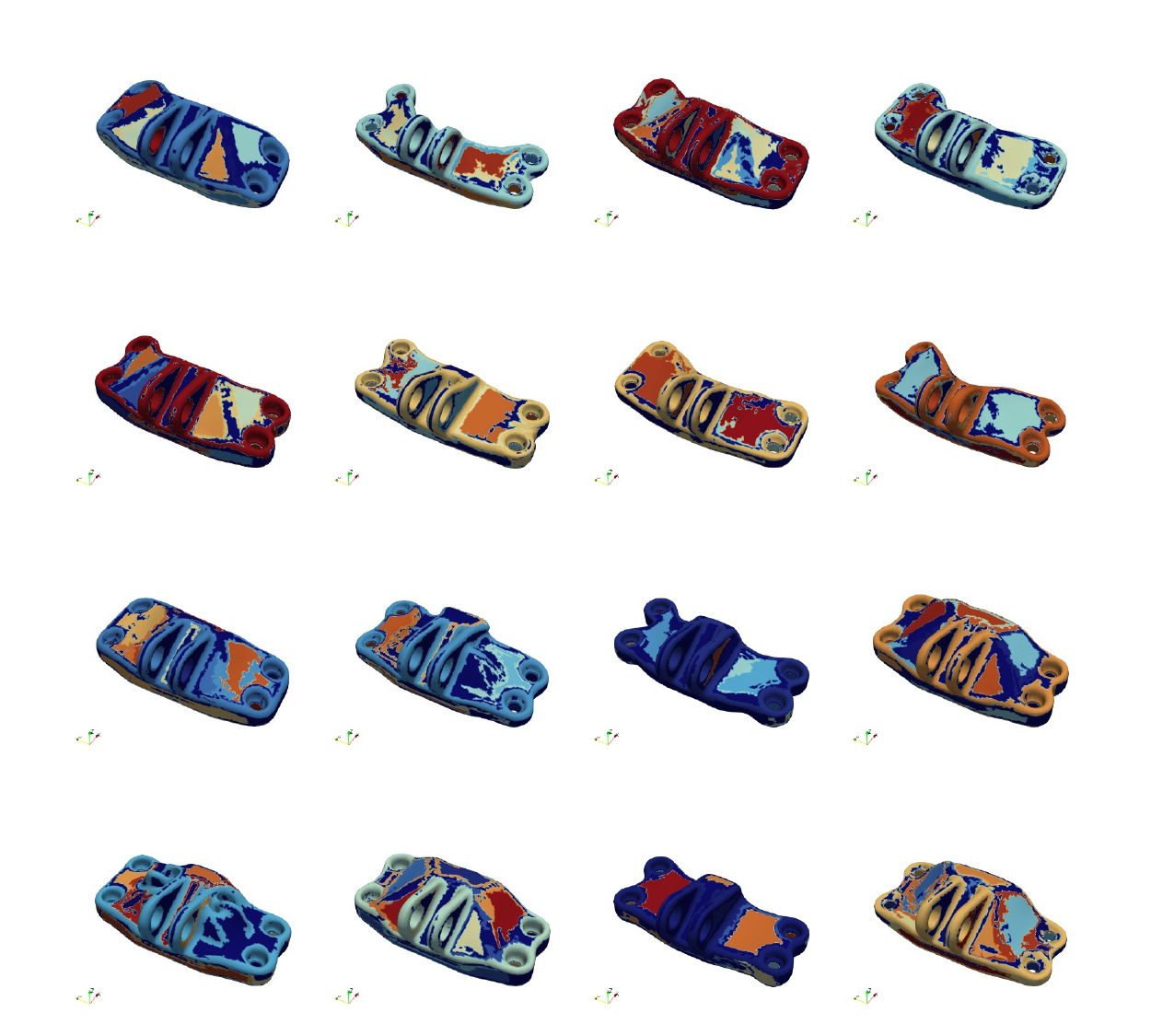}}
    \caption{The extracted tokens of different samples in JEB dataset are presented.}
    \label{fig.JEB_tokens}
  \end{center}
\end{figure}

\begin{figure}[H]
  \begin{center}
    \centerline{\includegraphics[width=0.55\columnwidth]{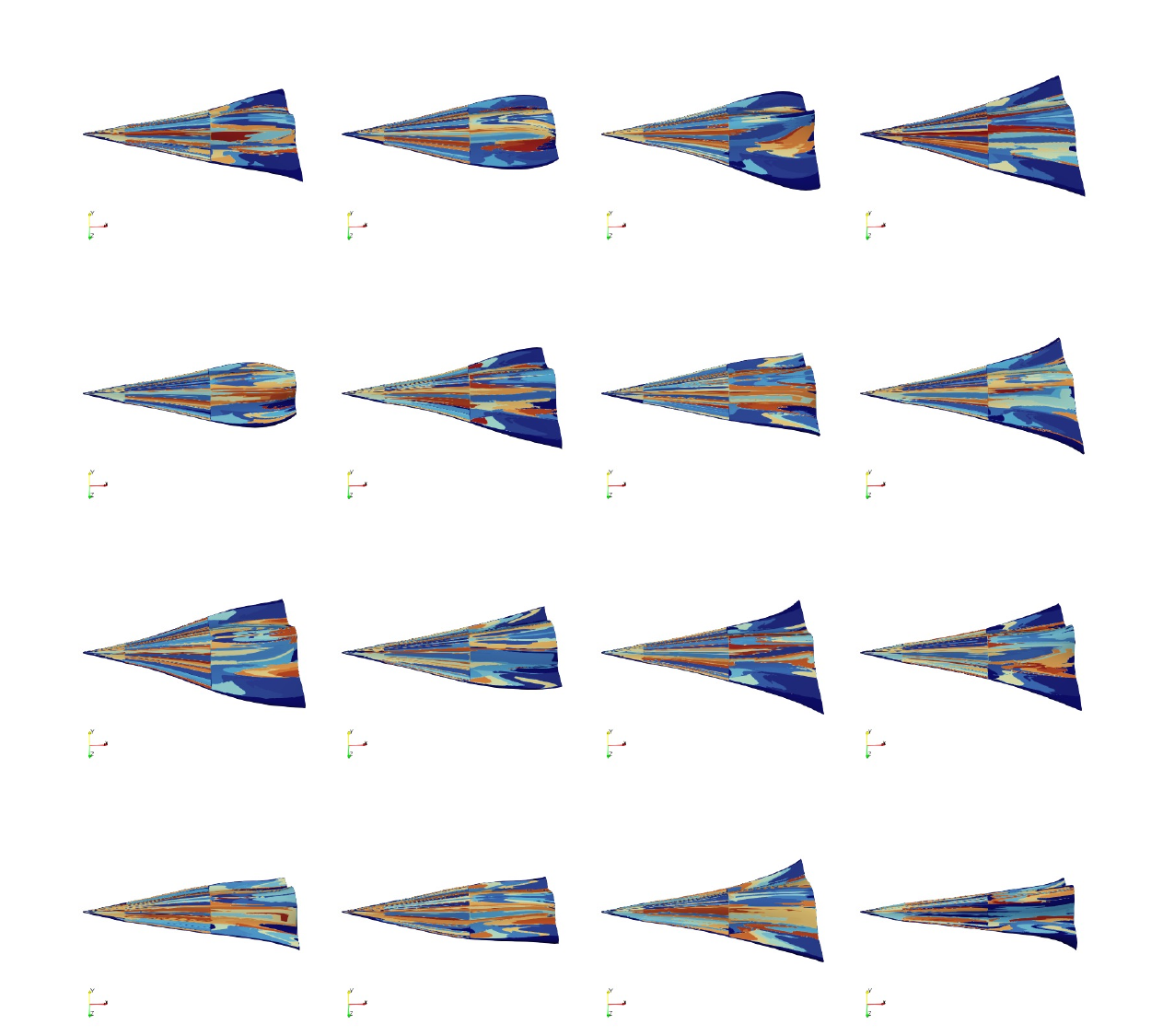}}
    \subcaption{Top view}
    \vspace{1em}
    \centerline{\includegraphics[width=0.55\columnwidth]{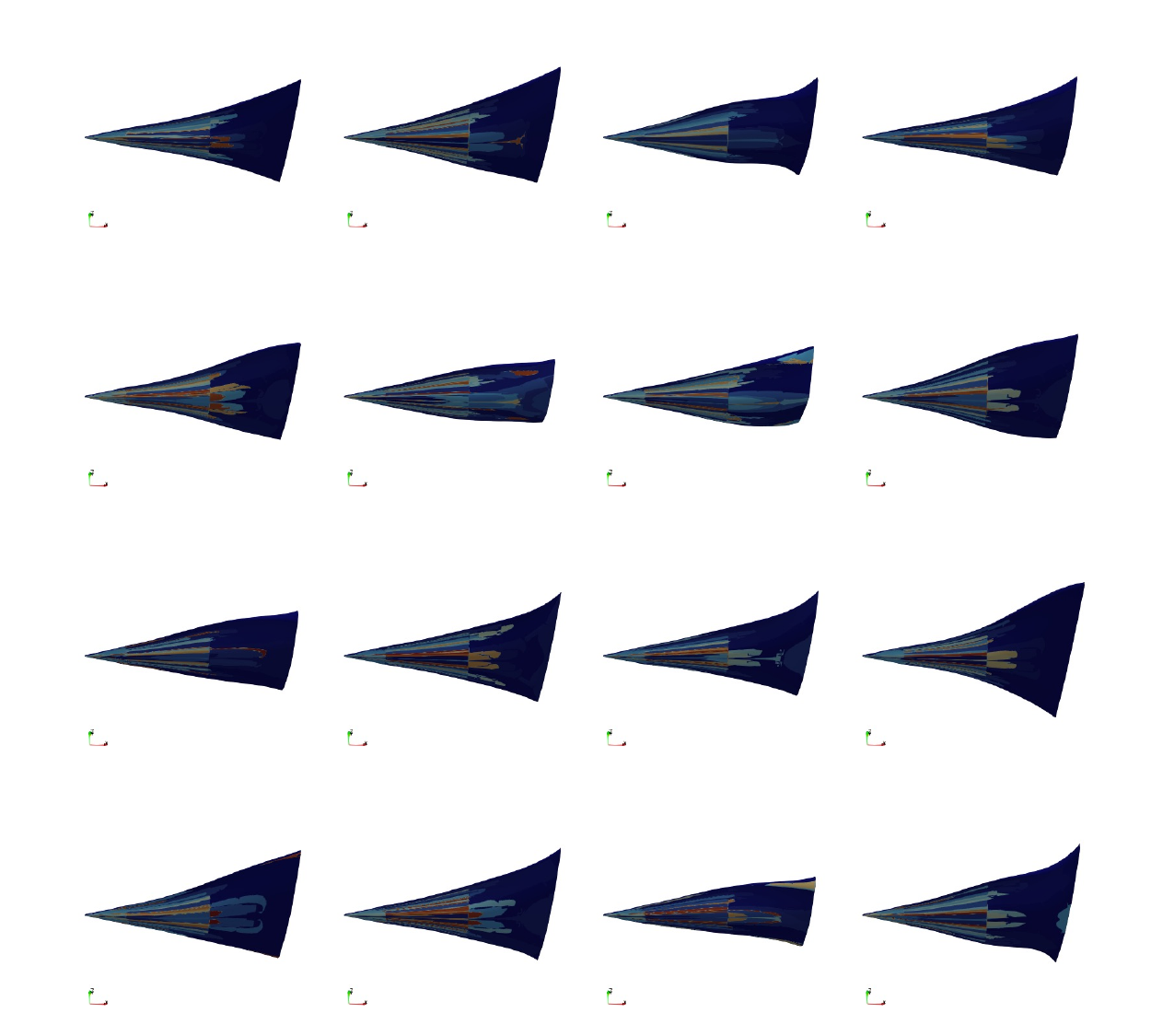}}
    \subcaption{Bottom view}
  \end{center}
  \caption{The extracted tokens of different samples in JEB dataset are presented.}
    \label{fig.Aircraft_tokens}
\end{figure}

\begin{figure}[H]
  \begin{center}
    \centerline{\includegraphics[width=0.55\columnwidth]{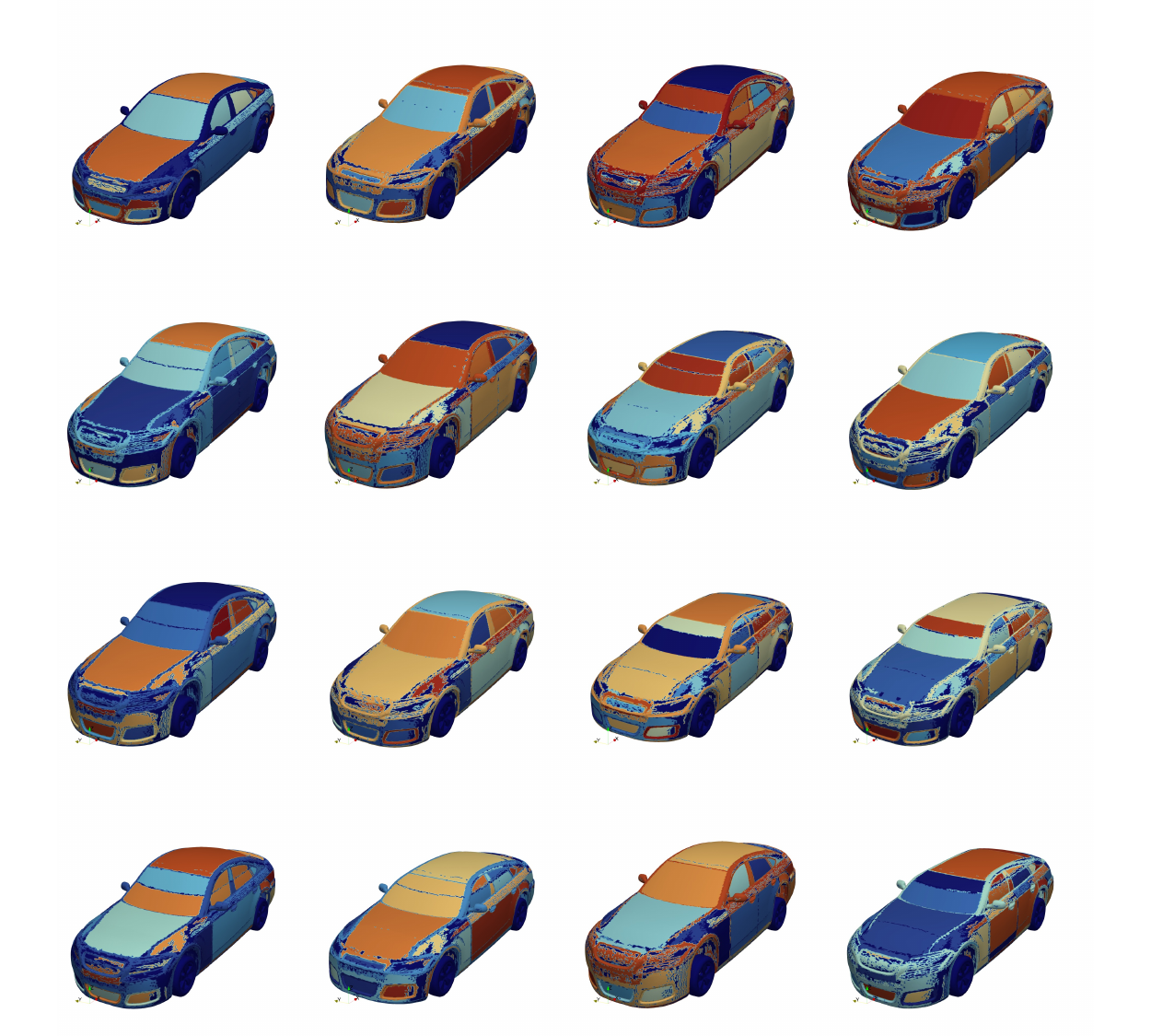}}
    \subcaption{Front view}
    \vspace{1em}
    \centerline{\includegraphics[width=0.55\columnwidth]{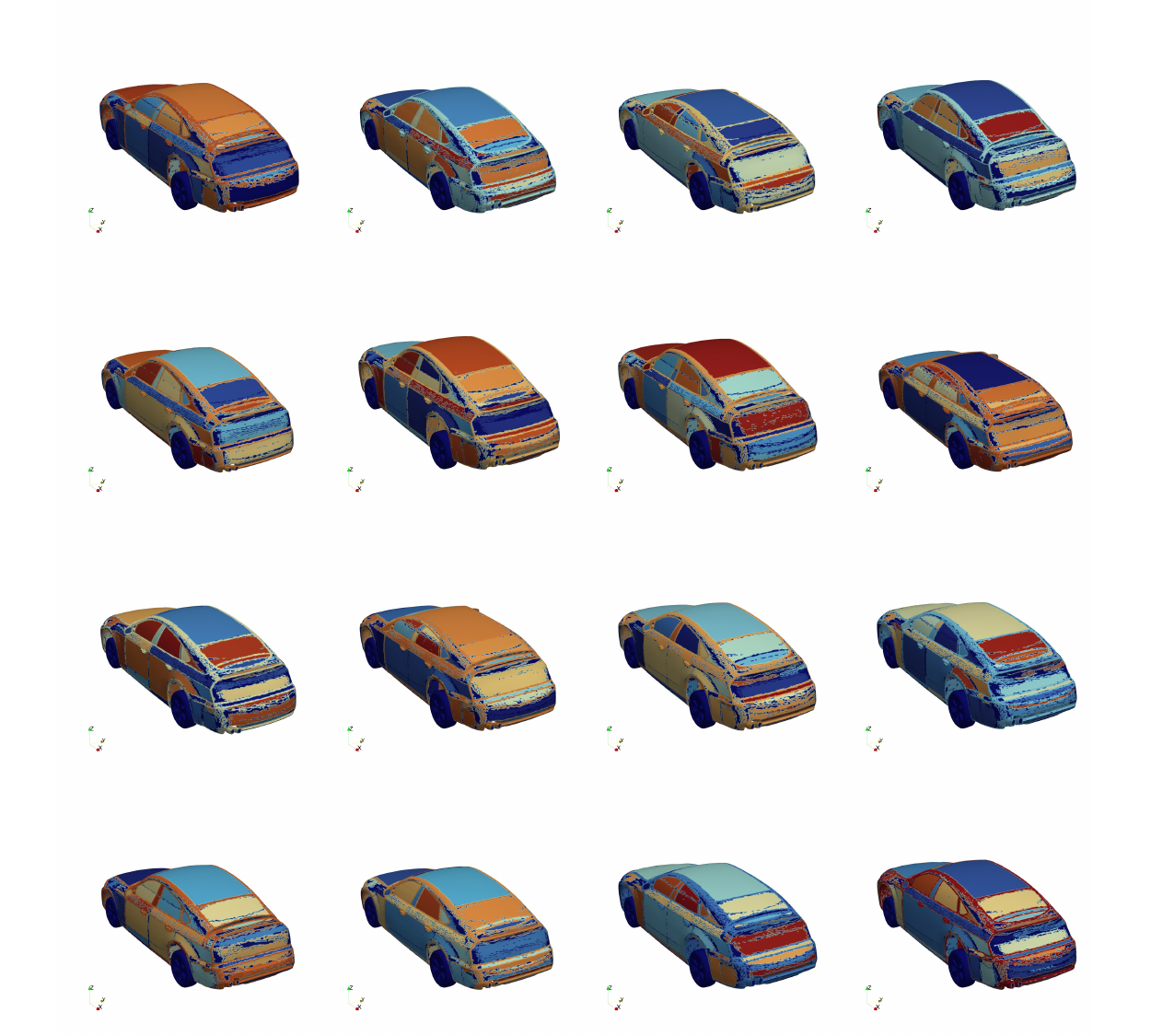}}
    \subcaption{Back view}
  \end{center}
  \caption{The extracted tokens of different samples in DrivAerNet++ dataset are presented.}
    \label{fig.drivAer_tokens}
\end{figure}

\begin{figure}[H]
  \begin{center}
    \centerline{\includegraphics[width=0.55\columnwidth]{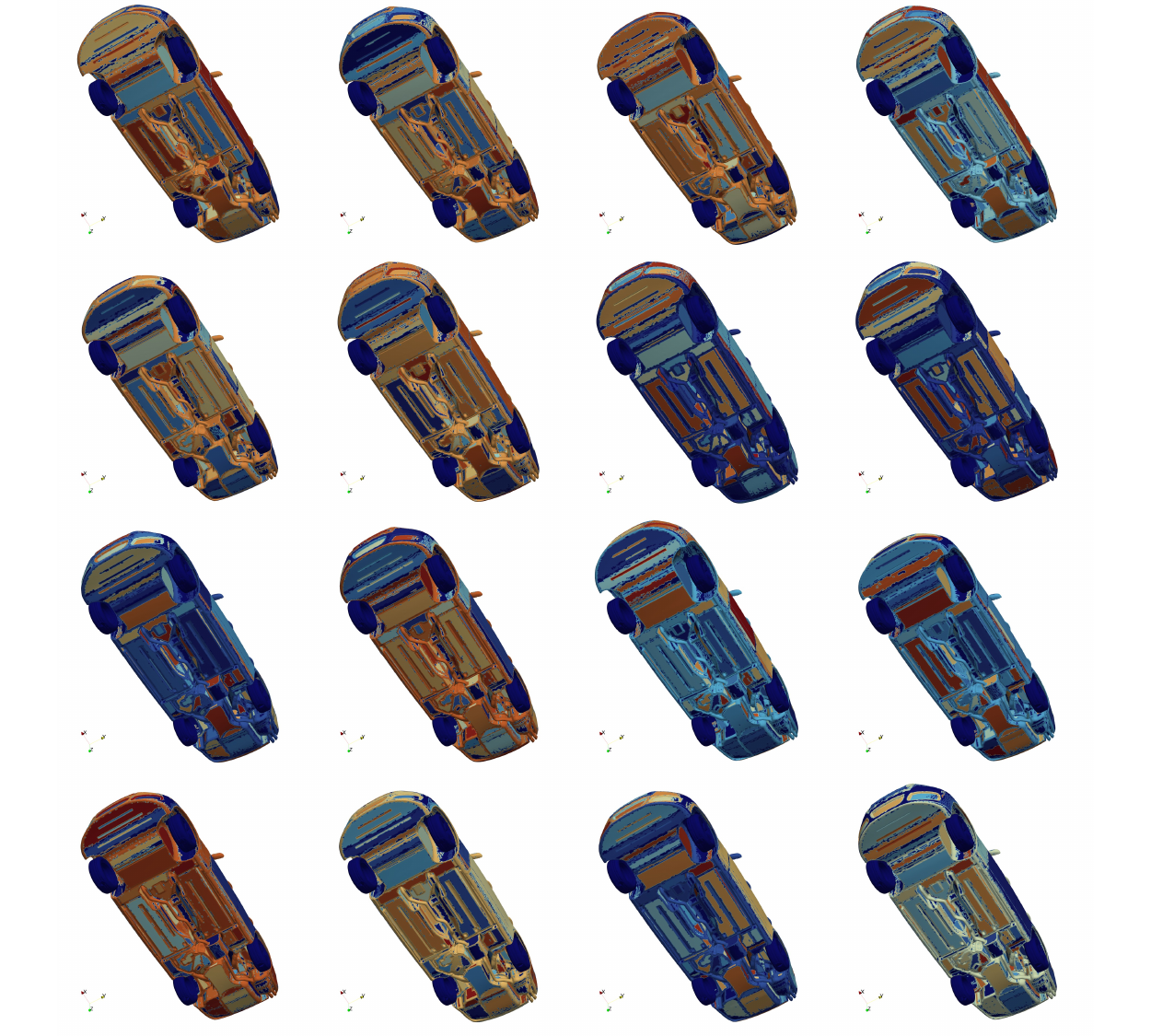}}
    \caption{The bottom view of the extracted tokens of different samples in drivAerNet++ dataset are presented.}
    \label{fig.drivAer_bottom_tokens}
  \end{center}
\end{figure}

\section{Implementation details and baseline comparison \label{App.baselines}}

To ensure a rigorous and fair comparison, all models are implemented within a unified experimental framework that standardizes architectural depth, hidden dimensions, and spatial resolution. Based on a combination of original implementations and GPU memory constraints, we set the default depth to 8 layers for point-based models and 4 layers for graph-based models. Hidden dimensions are assigned by model family: 128 for point-based networks (e.g., PointNet, Galerkin, GNOT, Transolver, Transolver++, and PGD-NO), 32 for graph-based operators (e.g., GNO, MeshGraphNet), and 16 for grid-based models (e.g., GINO). GELU is employed as the default activation function across all architectures. For point cloud inputs, graph connectivity is constructed using the KDTree algorithm, while grid-based models utilize a maximum hidden graph resolution of 100 voxels to maintain sufficient density.

To handle the extreme disparity in mesh scales—ranging from thousands to 2.5 million points—we implement a scale-aware adaptation strategy. For million-scale datasets such as DrivAerNet++ and Aircraft, where most baselines encounter memory limitations, we adopt a consistent subsampling approach. Inputs are reduced to 50k mesh points, with manifold structure preserved via K-Nearest Neighbors (KNN) reconstruction. During training, a random subsampling strategy is applied at the start of each epoch to increase data diversity while maintaining a constant step count. For evaluation, we aggregate multiple overlapping subsampled predictions into a single, high-fidelity reconstructed output. This ensures that even memory-constrained models can be evaluated on industry-scale geometries without compromising structural integrity.

To isolate architectural efficacy from optimization artifacts, all models are trained in a standardized environment using NVIDIA GH200 GPUs (100GB). We split each dataset into 60\% training, 10\% validation, and 30\% testing sets. Training is conducted for 200 epochs with a batch size of 1, utilizing the Adam optimizer \cite{adam} ($\beta_1 = 0.9, \beta_2 = 0.999$) and a weight decay of $10^{-4}$. The base learning rate of 0.001 is managed by a OneCycleLR scheduler to facilitate stable convergence. In instances where a configuration exceeds available GPU memory, we reduce the hidden dimension to the largest size that fits the constraint.

By unifying these training-side variables, we ensure that the reported $L_2$ relative errors and Quantities of Interest (QoI) reflect the intrinsic predictive power of the model architectures rather than variations in hyperparameter tuning. Each training step computes gradients based on a single graph to maintain consistency across different geometric complexities. This rigorous protocol allows us to highlight the performance gains specifically attributable to the structural design of PGD-NO.

\section{Metric \label{App.metric}}

To evaluate the predictive performance of our model across diverse physical domains, we employ both a global field reconstruction metric and task-specific engineering measurements. This dual-layered evaluation ensures that the model captures the continuous solution space of the governing Partial Differential Equations (PDEs) while maintaining precision in the critical scalar values used for industrial design.

The primary metric for quantifying prediction accuracy is the $L^2$ relative error over the entire field. Specifically, we predict temperature for the heat sink dataset, stress for the JEB dataset, pressure for the Aircraft and DrivAerNet++ datasets, and velocity in the X-direction for the CFD-VOL dataset. To validate the model for specific design tasks, we also evaluate several critical physical quantities. For the heat sink dataset, we extract the temperature distributions for both the chip and the heat sink. We specifically monitor the average chip temperature, as it serves as a direct indicator of heat retention and cooling efficiency. For the JEB dataset, we focus on the accuracy of the maximum predicted stress to ensure the design remains within safe operating limits. In aerodynamic applications, we evaluate the DrivAer dataset using the drag coefficient ($C_d$) to measure wind resistance. For the aircraft dataset, we calculate the lift coefficient ($C_l$) to determine flight viability. Finally, for the volumetric CFD data, a single scalar value is often insufficient to characterize the fluid pattern. Consequently, we evaluate the fluid field predictions on the middle cross-section ($v_{\text{mid}}$), as this plane contains the most significant information regarding flow structures and boundary layer interactions.

\subsection{Field prediction error}
To assess the global accuracy of the predicted physical fields across the entire spatial domain, we utilize the \textbf{Relative $L_2$ Error}. This metric provides a normalized measure of the discrepancy between the model prediction $\hat{y}$ and the ground-truth field $y$, allowing for consistent comparison across different physical units and scales. It is defined as:
\begin{equation}
    \text{Relative } L_2 = \frac{\|\hat{y} - y\|_2}{\|y\|_2} = \frac{\sqrt{\sum_{i=1}^{n} (\hat{y}_i - y_i)^2}}{\sqrt{\sum_{i=1}^{n} y_i^2}}.
\end{equation}
This error is computed individually for each state variable in the output vector. 

\subsection{Quantities of Interest (QoI)}
Beyond field-wide errors, we evaluate specific \textbf{Quantities of Interest (QoI)} that serve as the primary performance indicators in engineering design loops. These scalars are often derived through the integration or extremum searching of the predicted fields:

\begin{itemize}
    \item \textbf{Heat Sink (Average Chip Temperature $\bar{T}_{\text{chip}}$):} The critical thermal performance metric is the average temperature at the contact interface between the heat source and the cooling fins:
    \begin{equation}
        \bar{T}_{\text{chip}} = \frac{1}{A_{\text{base}}} \int_{A_{\text{base}}} T(\mathbf{x}) dA
    \end{equation}

    \item \textbf{JEB (Maximum von Mises Stress $\sigma_{\text{max}}$):} For structural safety, we identify the peak stress concentration within the geometry to predict potential material yielding or structural failure:
    \begin{equation}
        \sigma_{\text{max}} = \max_{\mathbf{x} \in \mathcal{M}} \sigma_{vm}(\mathbf{x})
    \end{equation}

    \item \textbf{DrivAer++ (Drag Coefficient $C_D$):} Automotive efficiency is measured by the drag coefficient, which requires integrating the predicted surface pressure $p$ and wall shear stress $\tau_w$ in the direction of the flow:
    \begin{equation}
        C_D = \frac{2}{\rho v^2 A} \int_A (p \mathbf{n} \cdot \mathbf{i} + \tau_w \mathbf{t} \cdot \mathbf{i}) dA
    \end{equation}

    \item \textbf{Aircraft (Lift Coefficient $C_L$):} Aerodynamic performance for flight stability is determined by the lift coefficient, derived by integrating the pressure distribution over the wing surface along the vertical axis $\mathbf{k}$:
    \begin{equation}
        C_L = \frac{2}{\rho v^2 A} \int_A (p \mathbf{n} \cdot \mathbf{k}) dA
    \end{equation}

    \item \textbf{CFD-VOL (Cross-Cut Velocity Error $\text{Err}_{\text{cut}}$):} For large-scale volumetric flow, we evaluate the relative $L_2$ error of the velocity vector field $\mathbf{u}$ along specific 2D planes to ensure accurate capture of internal wake structures:
    \begin{equation}
        \text{Err}_{\text{cut}} = \frac{\|\hat{\mathbf{u}}_{\text{plane}} - \mathbf{u}_{\text{plane}}\|_2}{\|\mathbf{u}_{\text{plane}}\|_2}
    \end{equation}
\end{itemize}

\section{Error map clarification \label{app.err_map}}

The visual analysis of the error distribution across diverse datasets highlights that the PGD-NO framework consistently achieves lower and more localized error patterns compared to the Transolver++ baseline. In thermal simulations like the Heat Sink, Transolver++ exhibits notable error clusters along the fin interfaces, whereas PGD-NO maintains a nearly uniform low-error profile across the entire geometry. This trend extends to complex structural and aerodynamic cases; for instance, in the JEB and Aircraft benchmarks, Transolver++ shows significant error concentrations in high-curvature regions and leading edges, while PGD-NO successfully mitigates these artifacts through its more effective geometric tokenization. In large-scale fluid dynamics simulations, such as the DrivAerNet++ surface and the corresponding velocity cross-sections, PGD-NO demonstrates superior precision in capturing boundary layer interactions and wake turbulences. By decoupling geometric feature extraction from point-wise querying, the model ensures that the prediction error remains minimal even in high-gradient areas, proving its robustness for high-fidelity industrial PDE modeling.

% \begin{figure}[ht]
%   \begin{center}
%     \centerline{\includegraphics[width=\columnwidth]{Figs/Main/error.pdf}}
%     \caption{Error map comparison between Transolver++ and PGD-NO. Geometric references on the left provide spatial context for the residuals. For the Heat Sink and CFD-VOL datasets, we visualize the direct physical fields due to their specific structural or external domain requirements. For all other benchmarks, error maps are overlaid onto the extracted geometry tokens to demonstrate localized model accuracy across distinct structural regions.}
%     \label{fig.error_plot_detail}
%   \end{center}
% \end{figure}

\end{document}